\documentclass{article}
   \usepackage[final,nonatbib]{neurips_2024}

\usepackage[utf8]{inputenc} % allow utf-8 input
\usepackage[T1]{fontenc}    % use 8-bit T1 fonts
\usepackage{hyperref}       % hyperlinks
\usepackage{url}            % simple URL typesetting
\usepackage{booktabs}       % professional-quality tables
\usepackage{amsfonts}       % blackboard math symbols
\usepackage{nicefrac}       % compact symbols for 1/2, etc.
\usepackage{microtype}      % microtypography
\usepackage{xcolor}         % colors
\usepackage{marvosym}
\usepackage{graphicx}
\usepackage{amsmath}
\usepackage{hyperref}
\usepackage{array}
\usepackage{caption}
\usepackage{wrapfig}        % half page image
\usepackage{multirow}       % multi row for table
\usepackage{color}          % color
\usepackage{diagbox}        %
\usepackage{enumitem}
\usepackage{footnote}

\definecolor{url_color}{RGB}{42, 83, 163}

\title{DiffPano: Scalable and Consistent Text to Panorama Generation with Spherical Epipolar-Aware Diffusion}

\vspace{-2.5em}
\author{
  Weicai Ye$^{1,3,}$\thanks{: Equal Contribution. \textrm{\Letter}: Corresponding Author. } \quad Chenhao Ji$^{2,*}$ \quad Zheng Chen$^{4}$ \quad Junyao Gao$^{2}$ \quad Xiaoshui Huang$^{3}$ \\ Song-Hai Zhang$^{4}$ \quad Wanli Ouyang$^{3}$ \quad Tong He$^{3,\textrm{\Letter}}$ \quad Cairong Zhao$^{2,\textrm{\Letter}}$ \quad Guofeng Zhang$^{1,\textrm{\Letter}}$
  \\
  $^1$State Key Lab of CAD\&CG, Zhejiang University \quad
  $^2$Tongji University\\
  $^3$Shanghai AI Laboratory
  $^4$Tsinghua Univerisity
  \\
  \texttt{\small maikeyeweicai@gmail.com \quad jichenhao@tongji.edu.cn \quad zhangguofeng@zju.edu.cn}
}

\begin{document}
\maketitle
\begin{figure}[ht]
    \begin{center}
    % \vspace{-3.3em}
    \vspace{-3em}
\centerline{\includegraphics[width=1\linewidth]{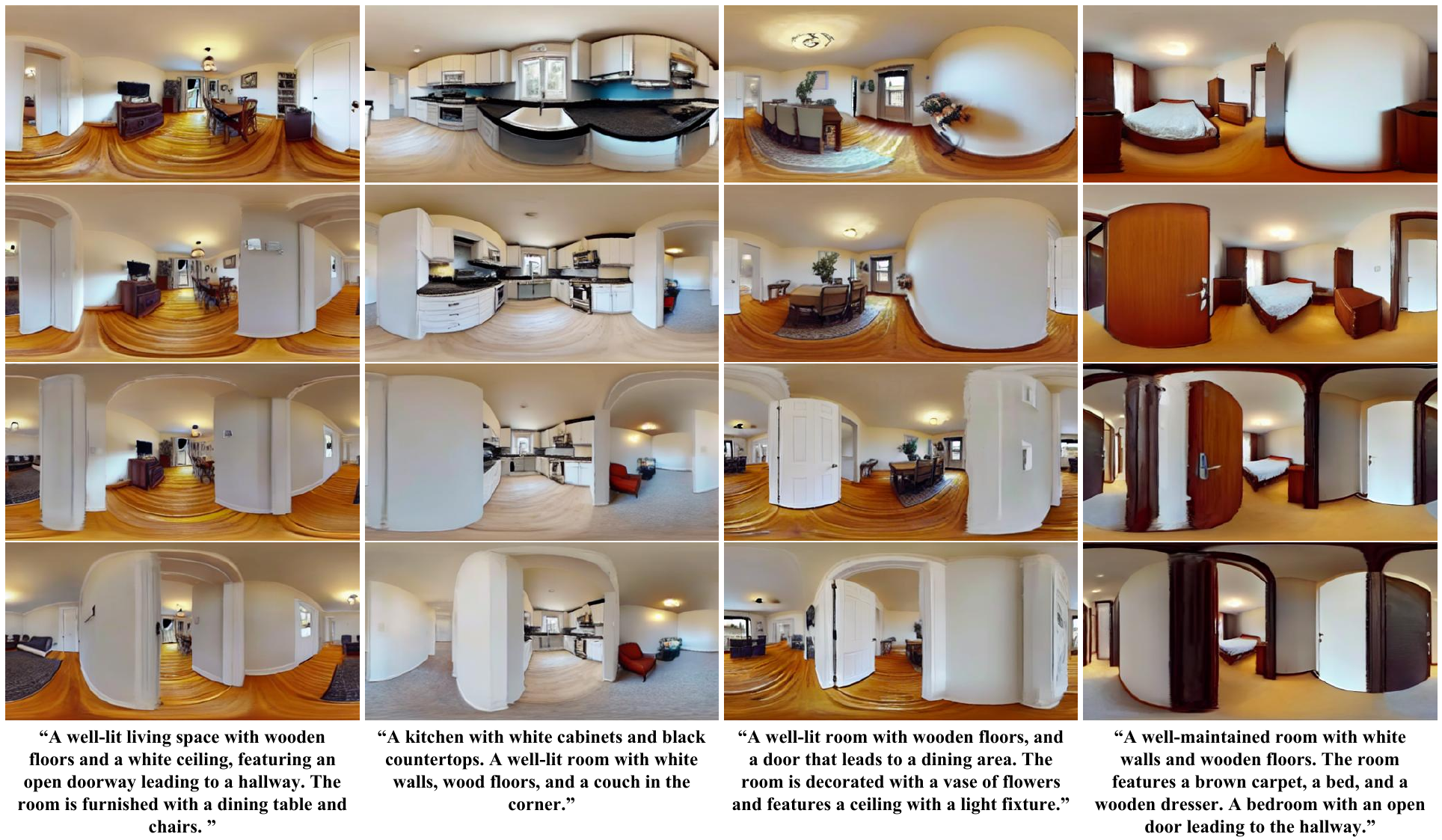}}
        % \vspace{-0.5em}
        \vspace{-0.3em}
        \caption{\textbf{DiffPano allows scalable and consistent panorama generation (i.e. room switching) with given unseen text descriptions and camera poses.} Each column represents the generated multi-view panoramas, switching from one room to another.}
        \label{fig:teaser}
    \end{center}
    % \vspace{-2.5em}
\end{figure}

\begin{abstract}
Diffusion-based methods have achieved remarkable achievements in 2D image or 3D object generation, however, the generation of 3D scenes and even $360^{\circ}$ images remains constrained, due to the limited number of scene datasets, the complexity of 3D scenes themselves, and the difficulty of generating consistent multi-view images. To address these issues, we first establish a large-scale panoramic video-text dataset containing millions of consecutive panoramic keyframes with corresponding panoramic depths, camera poses, and text descriptions. Then, we propose a novel text-driven panoramic generation framework, termed DiffPano, to achieve scalable, consistent, and diverse panoramic scene generation. Specifically, benefiting from the powerful generative capabilities of stable diffusion, we fine-tune a single-view text-to-panorama diffusion model with LoRA on the established panoramic video-text dataset. We further design a spherical epipolar-aware multi-view diffusion model to ensure the multi-view consistency of the generated panoramic images. Extensive experiments demonstrate that DiffPano can generate scalable, consistent, and diverse panoramic images with given unseen text descriptions and camera poses. Code and dataset are available at \url{https://zju3dv.github.io/DiffPano}.

% on the built panorama datase 
% HM3D, MP3D, and Replica 
% To achieve scalable panorama generation,  propose a keyframe-based condition to generate longer video frames.
% and our dataset can improve the performance of panoramic radiance fields. 
% Our DiffPano can support a range of compelling real-world applications: such as panorama to 3D, text to 3D panorama, and 3D panorama stylization.
\end{abstract}

\section{Introduction}
Generating scenarios from the text descriptions that meet one's expectations is an imaginative and marvelous journey, which has many potential applications, such as VR roaming~\cite{yang2024dreamspace} for metaverse~\cite{ye2021superplane, yu2022improving, yu2020learning, zhang2021arcargo}, physical world simulation~\cite{videoworldsimulators2024, chen2024gst, gao2024luminat2x}, embodied agents in scene navigation and manipulation~\cite{zhu2024point}, etc. With the advent of the AIGC era, several works~\cite{chen2024meshanything, gao2024luminat2x, epidiff, zero123, zero123++, mvdream, mvdiffusion, pose_guided, geco} on object generation even scene generation has emerged. However, they generally only generate a series of perspective images, making it impossible to comprehensively simulate the entire environment for scene understanding~\cite{liu2024point, huang2024nerfdet++, ye2022hybrid} and reconstruction~\cite{cai2024dynasurfgs, chen2024pgsr, ye2023PVO, Ye2023IntrinsicNeRF, liu2021coxgraph, ming2022idf, ming2024viper, liu2024difflow3d, li2020saliency, ye2024DATAP-SfM, chen2024gigags, cui2024streetsurfgs, tang2024hisplat, tang2024nd,  wang2024neurodin, ye2024FedSurfGS}. Given this, some methods~\cite{panfusion, text2light} try to generate panoramas to solve these problems by taking advantage of the inherent characteristics of panoramic images, which can capture the surrounding environment with a single shot. 

These methods can be roughly divided into four categories: 1) Directly single-view equirectangular projection (ERP) panorama generation~\cite{panfusion}. However, its camera is immovable, making it incapable of scene exploration; 2) Multiple perspective views generation methods~\cite{epidiff, mvdream, mvdiffusion, zeronvs} without considering multi-view panorama generation. 3) The inpainting solutions are based on the infinite expansion of a single perspective view, which lacks 3D awareness. Single scene optimization methods~\cite{perf2023} with inpainting have no generalization ability and cost too much time for optimization. 4) Directly extending the multiple perspective views generation methods~\cite{mvdream} to the ERP panorama, which is difficult to converge and results in poor multi-view consistency (see Fig.~\ref{fig:cmp_mvdream}). 
% Although these methods achieve promising results, they can only generate one panoramic image at a time, and their cameras are immovable, making them incapable of scene exploration and not conducive for AI agents to perform practical tasks. 

This paper aims to generate scalable and multi-view consistent panoramic images from text descriptions and camera poses (see Fig.~\ref{fig:teaser}) with many potential applications such as immersive VR roaming with unlimited scapes and preview for interior home design.  
However, achieving this goal is not trivial. To the best of our knowledge, there is currently a lack of rich and diverse panoramic datasets to meet the task of text-to-multi-view ERP panorama generation. To this end, we propose a novel panoramic video-text dataset and a generation framework suitable for the text-to-multi-view panorama generation task, advancing the development of this field. Specifically, we first establish a large-scale panoramic video-text dataset using Habitat Simulator~\cite{habitat} (see Sec.~\ref{Pano Video-Text Dataset}), which contains millions of panoramic keyframes and corresponding panoramic depths, camera poses, and text descriptions. Next, built upon the proposed dataset, we propose a generation framework for consistent multi-view ERP panorama generation (see Sec.~\ref{diffpano}), termed DiffPano. The DiffPano framework consists of a single-view text-to-panorama diffusion model (see Sec.~\ref{pano_sd}) and a spherical epipolar-aware multi-view diffusion model (see Sec.~\ref{multi}). The single-view text-to-panorama diffusion model is obtained by fine-tuning the stable diffusion model~\cite{stable_diffusion} of perspective images using LoRA~\cite{lora}. Considering that the single-view pano-based diffusion model cannot guarantee the consistency of generated multi-view panoramas with different camera poses, we derive a spherical epipolar constraint applicable to panoramic images, inspired by the perspective epipolar constraint. We then incorporated it as a spherical epipolar-aware attention module (see Sec.~\ref{Spherical epipolar attention}) into the multi-view panoramic diffusion model to ensure the multi-view consistency of the generated ERP panoramic images.

% Considering that the single-view pano-based diffusion model cannot guarantee the consistency of the generated multi-view panoramic images with different camera poses, inspired by the perspective epipolar constraint, we derive a spherical epipolar constraint applicable to panoramic images and incorporate it into the multi-view panoramic diffusion model as a spherical epipolar attention module (see Sec.~\ref{Spherical epipolar attention}) to ensure the multi-view consistency of the generated ERP panoramic images.

% To better verify our method, we propose some metrics for multi-view ERP panorama generation, such as the quality of multi-view panoramic images, multi-view consistency, the consistency between the text and multi-view panoramas. 

Since there are no related methods for comparison, we try to extend the MVDream method~\cite{mvdream} to generate multi-view ERP panoramas. Trained and tested on the proposed panoramic video-text dataset, extensive experiments demonstrated that compared to the modified MVDream, our proposed multi-view panorama generation based on spherical epipolar-aware attention can generate more scalable and consistent panoramic images. Our method also demonstrates the generalization ability of the original diffusion model to generate satisfactory multi-view ERP panoramas with given unseen text descriptions and camera poses. 

Our contributions can be summarized as follows: 1) To the best of our knowledge, we are the first to propose a scalable and consistent multi-view panorama generation task from text descriptions and camera poses. 2) We established a large-scale diverse and rich panoramic video-text dataset, which fosters the research of text-to-panoramic video generation. 3) We propose a novel text-driven panoramic generation framework with a spherical epipolar attention module, allowing scalable and consistent panorama generation with unseen text descriptions and camera poses.

\section{Related work}
% In this section, we briefly review the methods of single-view panorama generation in Section~\ref{singleview_works} and multi-view perspective image generation in Section~\ref{multiview_works}. Then we review the existing panorama datasets in Section~\ref{panorama_dataset}.

% \subsection{Diffusion model}

% \czcomment{delete this part: Diffusion Model

% % 自从扩散模型发布之后，生成式AI就进入了飞速发展时期，特别是在图像生成领域。隐式扩散模型的出现解决了扩散模型采样速度过慢的问题，它通过将扩散模型压缩到潜空间中来提高采样效率并减少计算消耗。先前的工作表明隐式扩散模型在图像生成中具有多样性和保持高质量的能力。为了将隐式扩散模型的图像生成能力运用到更多特定任务中，许多方法通过让模型学习图像的风格并生成相应风格的图像。我们在隐式扩散模型的基础上探索了它同时生成多视角全景图像的能力。

% Since the introduction of diffusion models, generative AI has entered a period of rapid development, particularly in the field of image generation. The emergence of latent diffusion models (LDMs) has solved the problem of slow sampling speed in traditional diffusion models. Latent diffusion models achieve higher sampling efficiency and reduced computational cost by compressing the diffusion model into the latent space. Previous work has demonstrated the ability of LDMs to generate diverse and high-quality images. To apply the image generation capabilities of LDMs to more specific tasks, many approaches have been proposed to enable the model to learn and generate images in various styles. Building upon LDMs, we explore their ability to simultaneously generate multi-view panoramic images.
% }

\subsection{Single-View Panorama Generation}
\label{singleview_works}
% 随着隐式扩散模型在全景图像生成领域的能力逐渐被挖掘，最近有许多基于隐式扩散模型的全景图像生成方法取得了非常惊艳的效果。MVDiffusion通过多视角Correspondence-Aware扩散模型来同时生成8张固定视角的透视图，并对透视图进行拼接来生成全景图，但是最终生成的全景图像更像一张范围广阔的广角图像。为了提高生成的全景图的左右边缘连续性，这也是全景图像的本质属性，也有很多工作在这一方面进行了探索。这些方法尝试在生成全景图的过程中让图像的左右边缘进行交互，从而在一定程度上提高了生成的全景图像的左右边缘连续性。全景图像生成领域中另一个关键问题是全景图像与透视图像之间的域差异，这一直是一个长期存在的挑战。为了解决这个问题, PanFusion 提出了一种新颖的双分支扩散模型, 减轻了透视图投影到全景图中的失真的同时提供全局布局指导。然而, PanFusion 因为数据集原因，生成的全景图顶部和底部都是模糊的，同时因为引入了更加复杂的模型结构，在推理生成全景图的过程中会花费更多的时间。为了兼顾计算速度和保证全景图像左右连续性，我们提出的基于全景图的Stable Diffusion只需要通过LoRA微调的方法就能让扩散模型学习到全景图的风格，在保证高生成速度的同时具备不错的左右边缘连续性。

Recently, latent diffusion model (LDM) methods have attracted widespread attention, and many single-view panorama generation works~\cite{panfusion, mvdiffusion, zhou2024holodreamer, li20244k4dgen, wu2023ipo, wang2023360, lu2024autoregressive, wang2024customizing, yang2024dreamspace} have emerged, achieving remarkably impressive results. Among them, MVDiffusion~\cite{mvdiffusion} simultaneously generates eight fixed-viewpoint perspective images through a multi-view Correspondence-Aware diffusion model and stitches them together to produce a panorama. However, it cannot support the generation of top and bottom views, and the generated panorama resembles wide-angle images with an extensive field of view rather than true $360^{\circ}$ images. Some methods~\cite{customizing360,diffusion360} solve this problem using equirectangular projection (ERP) and try to facilitate the interaction between the left and right sides during the panorama generation process to enhance the left-right continuity property inherent in ERP images. To address the domain gap between panorama and perspective images, PanFusion~\cite{panfusion} proposed a novel dual-branch diffusion model that mitigates the distortion of perspective images projected on panoramas while providing global layout guidance. However, its more complex model architecture incurs longer inference times for panorama generation. In addition, PanFusion cannot be expanded as an effective pre-trained model to the multi-view panorama generation task due to its excessive network parameters. To strike a balance between computational complexity and ensuring left-right continuity of panoramas, our proposed single-view panorama-based stable diffusion model only requires fine-tuning with LoRA~\cite{lora} to learn panoramic styles and achieve good edge continuity while maintaining higher generation speed and simpler architecture.

The existing single-view panorama generation methods cannot achieve scalable panorama generation. The core of our paper lies in the generation of multi-view consistent panoramic images, which we will introduce in Section~\ref{multiview_works}. More importantly, the single-view panoramic image generated by previous methods mainly supports 3DoF roaming, while our method can generate multi-view panoramic images for 6DoF roaming, which can serve as the inputs for $360^{\circ}$ Gaussian Splatting~\cite{omnigs} or $360^{\circ}$ NeRF~\cite{panogrf,structnerf}. Our method also has a great potential value in $360^{\circ}$ relightable novel view synthesis with the combination of $360^{\circ}$ multi-view inverse rendering method~\cite{mv_inverse}.

% \subsection{Diffusion-based multi-view synthesis}
\subsection{Multi-View Image Generation}
\label{multiview_works}
% 多视角生成任务可以被视为多个视角的新视角生成任务。在先前的新视角生成相关工作中，是将其中一个视角的图像作为条件并通过扩散模型生成另一视角下的图像。Zero123为后续许多针对物体的新视角生成任务打下了坚实的基础，pose-guided diffusion model探索了基于场景的新视角图像生成。上述方法在新视角生成任务中取得了很好的效果，而在多视角图像生成任务中，如果通过多次使用使用上述方法生成不同视角的图像，会导致多视角图像之间一致性较差。为了能够同时生成高质量的多视角图像，这些工作将扩散模型中的Unet改为多分支的形式，并通过不同分支图像之间的交互，达到同时生成多视角相互关联的图像的效果。为了实现多视角图像之间的交互，大部分方法通过引入相机视角来控制图像的生成，通过给定每一张生成图像的视角来让模型理解3D空间知识。目前大多数多视角生成任务是针对单一物体的多视角图像生成，或者是多视角场景透视图像的生成，基于全景图的多视角场景图像生成工作非常非常少。因为全景图与透视图之间的域差异，要实现多视角全景图的一致性更为复杂。我们提出的方法通过相机位姿在多张全景图像之间构造对极约束来实现跨视角交互，能够同时生成4张具有高度一致性且符合对应文本描述的全景图。

To the best of our knowledge, there is no work focusing on multi-view panorama generation. We review the existing works about multi-view generation for perspective images in this part.

% Zero123~\cite{zero123} laid a foundation for the multi-view generation task targeting objects, while the pose-guided diffusion model~\cite{pose_guided, zeronvs} explored novel view synthtsis focusing on scenes. However, applying them multiple times to generate different views in multi-view image generation tasks can lead to poor consistency among the multi-view images. To simultaneously generate high-quality multi-view images, these works~\cite{syncdreamer, mvdream, imagedream, wonder3d} modify the UNets in the diffusion model into a multi-branch form and achieve the effect of generating consistent multi-view images through the interaction between different branch. 
Zero123~\cite{zero123} laid the foundation for 3D object generation based on multi-view generation, while the pose-guided diffusion model~\cite{pose_guided, zeronvs} explored consistent view synthesis of scenes. However, iteratively applying the diffusion model to generate individual views in the multi-view generation task may lead to poor multi-view consistency of generated images due to accumulated errors. 
To generate high-quality multi-view images simultaneously, some methods~\cite{syncdreamer, mvdream, imagedream, Wonder3d} modify the UNets in the diffusion model into a multi-branch form and achieve the effect of generating consistent multi-view images through the interaction between different branch. 

Currently, most multi-view generation tasks focus on generating multi-view perspective images of single objects~\cite{mvdream, xu2024grm, yang2024consistnet, liu2024one, tang2023dreamgaussian, tang2025lgm} or scenes~\cite{pose_guided, gao2024cat3d}, while minimal research has been conducted on multi-view panorama generation. Narrow FoV (field-of-view) drawbacks of perspective images lead to the fact that the existing generation methods can only generate a very local region of the scene at a time. Our work focuses on the task of exploring the generation of $360^{\circ}$ images from multiple different viewpoints. Due to the camera projection difference between panoramic and perspective images, achieving consistency in multi-view panoramas is challenging. It is impossible to directly apply the existing epipolar attention module~\cite{pose_guided, epidiff} to multi-view panoramas. We strive to derive the spherical epipolar line formula for panoramic images and propose a spherical epipolar attention module to ensure the multi-view consistency of the generated panoramas.

\begin{figure}[ht]
    \begin{center}
    \vspace{-1em}
        \centerline{\includegraphics[width=1\linewidth]{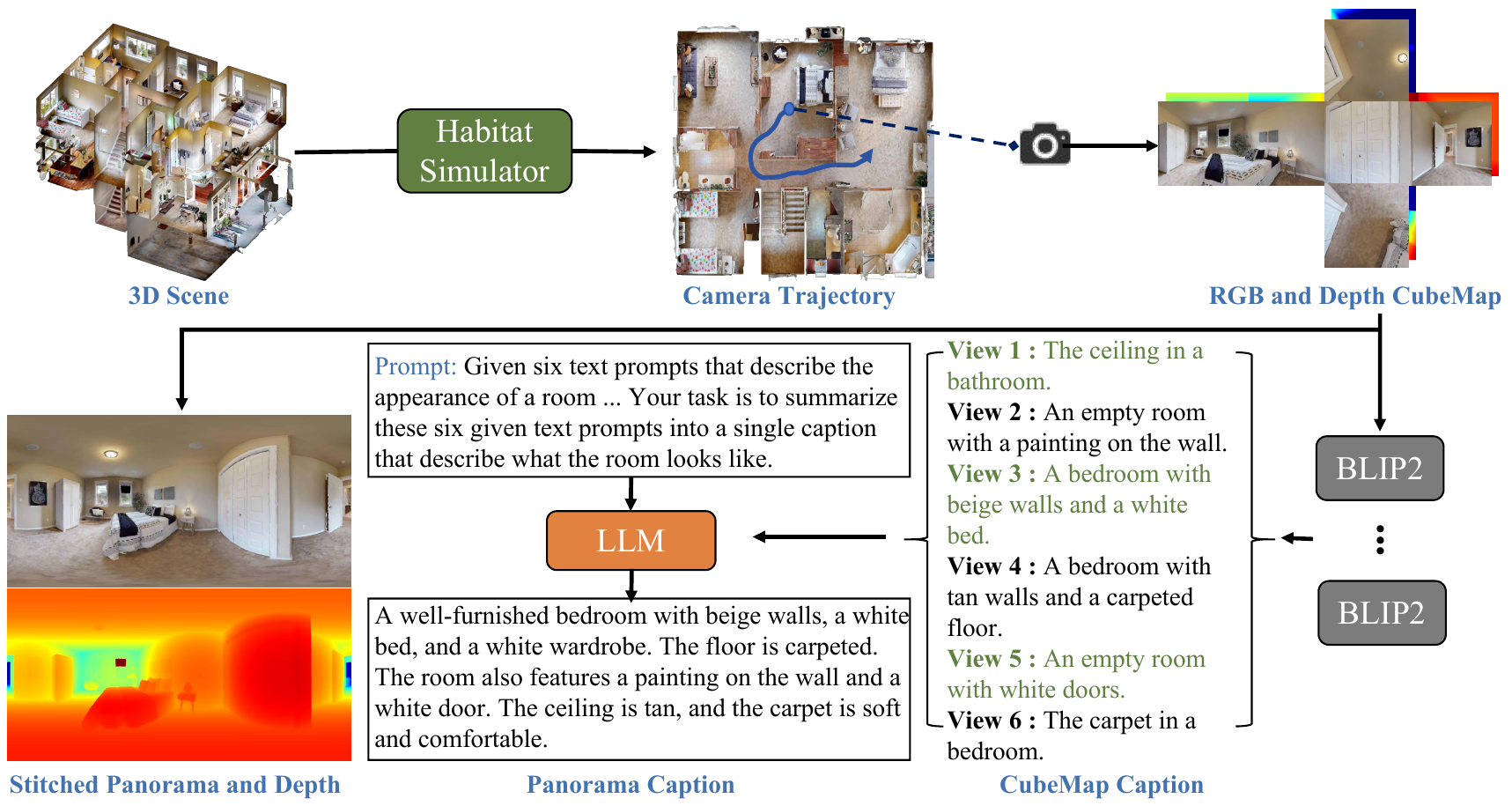}}
        \vspace{-0.1cm}
        \caption{\textbf{Panoramic Video Construction and Caption Pipeline.}}
        % \label{fig:stage1}
    \end{center}
    \vspace{-0.6cm}
\end{figure}

\subsection{Panoramic Dataset}
\label{panorama_dataset}
% 数据集的缺乏一直是限制全景图生成领域工作的重要原因之一。Structured3D包含了3.5K个房屋设计，以及与之对应的高清2D全景图像。但是我们发现其中部分全景图像的左右边缘存在光线亮度不一致的情况，这会导致利用该数据集训练的模型生成的全景图也带有这种现象，影响了全景图的质量。PanoFusion和MVDiffusion使用Matterport 3D中的skybox图像构造数据集，但是因为每一组skybox图像的顶上视角和底下视角图像的中间存在模糊的情况，最终拼接获得的全景图只有水平视角上是清晰的。并且以上全景数据集中的图像都没有对应的文本描述，不能直接用于文本到全景图像的生成任务。MVDiffusion利用blip2对Matterport 3D skybox中的透视图像生成对应文本描述，PanFusion则通过blip2直接对拼接成的全景图像生成非常简短的文本描述。这两种方式得到的全景图像对应文本描述存在不连贯和不精确的问题。此外，要生成多视角的全景图像我们还需要每张图像对应的相机位姿。为了解决数据集缺乏的问题，我们通过Habitat Simulator在Habitat Matterpoet 3D和Matterport 3D数据集中的每个场景对应的3D mesh中进行渲染，获取当前视角下的立方体六面透视图，并通过插值将其拼接成全景图像。通过使用blip2分别对立方体六面透视生成文本描述，再用llama2进行总结的方式生成全景图对应的完整文本描述。这样我们就获得了一个同时包含相机位姿、对应全景图以及全景图对应文本描述的数据集，便于进行后续的多视角全景图生成任务。

Great progress in text to single-view panorama generation has been witnessed. However, text-to-multi-view panorama generation is still a blank slate. One of the main limitations of this task is the lack of suitable datasets. The common panoramic datasets used in single-view panorama generation consist of indoor HDR dataset~\cite{gardner2017learning}, outdoor HDR dataset~\cite{zhang2017learning}, HDR360-UHD dataset~\cite{text2light}, Structured3D~\cite{structured3d}, Standford 2D-3D-S~\cite{armeni2017joint}, and Matterport3D dataset~\cite{matterport3d}, etc. Most of these datasets are relatively small in scale and only have single-view panoramas, which cannot support multi-view panorama generation, except Matterport3D~\cite{matterport3d}. In addition, the sky box images in Matterport3D~\cite{matterport3d} contain only sparse views. Although HM3D~\cite{hm3d} provides the textured mesh of 1000 scenes, it lacks the corresponding text description for each view. To generate multi-view panoramas, we render cube maps at each viewpoint in the 3D meshes of HM3D, using the Habitat Simulator~\cite{habitat}, and stitch them into panoramas. We generate complete text descriptions corresponding to the panoramas by using Blip2~\cite{blip2} to create text descriptions for each face of the cube map separately, and then summarizing them using Llama2~\cite{llama2}. In this way, we obtain a panoramic video-text dataset that includes camera poses, corresponding panoramas, and text descriptions of the panoramas, which facilitates subsequent multi-view panorama generation tasks.

\section{Panoramic Video-Text Dataset}
\label{Pano Video-Text Dataset}
% 因为目前没有可用的非常高质量的Panomara-text dataset，大部分文本到全景图像的生成任务都需要自己构建数据集。PanFusion中构造的数据集存在全景图顶部和底部模糊的情况，且对应的文本描述不够精确。我们利用Habitat Simulator在Habitat Matterport 3D（HM3D）数据集的场景中随机选取位置渲染立方体六面贴图，并将其通过差值拼接成全景图。这样得到的全景图虽然在图像质量上没有Matterport 3D skybox images拼接起来的好，但是我们可以得到顶部和底部都清晰的全景图像。为了给全景图像生成更加精确的文本描述，我们先对获取的立方体六面贴图分别用blip2生成对应的文本描述，再利用llama2进行总结，获取精确完整的文本描述。利用Habitat Simulator还可以在HM3D场景中根据相机轨迹渲染图像，生成同时包含相机位姿、全景图和对应文本描述的数据集，这将在3.2中的任务中使用。
Due to the lack of high-quality panorama-text datasets, most text-to-panorama generation tasks require researchers to construct their own datasets. The dataset constructed in PanFusion~\cite{panfusion} suffers from blurriness at the top and bottom of the panoramic images, and the corresponding text descriptions are not precise enough. To address these issues, we utilize the Habitat Simulator~\cite{habitat} to randomly select positions within the scenes of the Habitat Matterport 3D (HM3D)~\cite{hm3d} dataset and render the six-face cube maps. These cube maps are then interpolated and stitched together to form panoramas so we can obtain panoramas with clear tops and bottoms. To generate more precise text descriptions for the panoramas, we first use BLIP2~\cite{blip2} to generate corresponding text descriptions for each obtained cube map, and then employ Llama2~\cite{llama2} to summarize and receive accurate and complete text descriptions. Furthermore, the Habitat Simulator allows us to render images based on camera trajectories within the HM3D scenes, enabling the creation of a dataset that simultaneously includes camera poses, panoramas, and corresponding text descriptions. This dataset will be utilized in the multi-view panorama generation (see Sec.~\ref{multi}).

\begin{figure}[ht]
    \begin{center}
        \centerline{\includegraphics[width=1\linewidth]{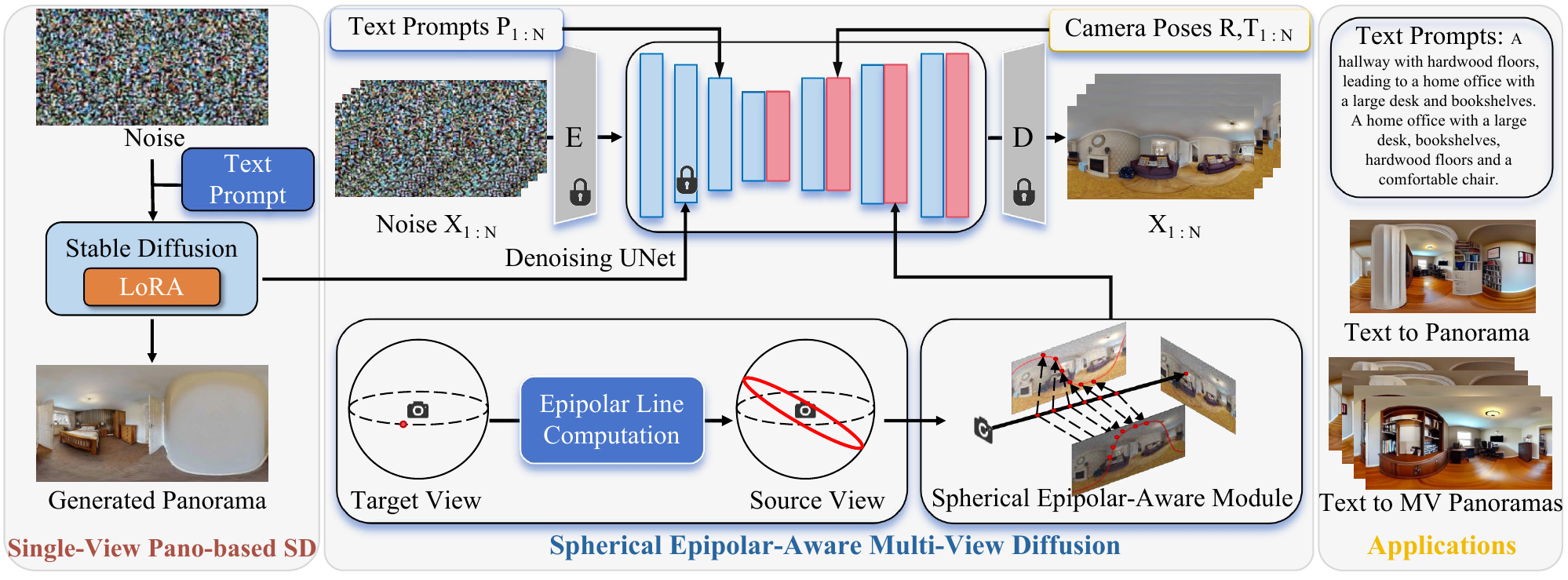}}
        \vspace{-0.1cm}
        \caption{\textbf{DiffPano Framework.} The DiffPano framework consists of a single-view text-to-panorama diffusion model and a spherical epipolar-aware multi-view diffusion model. It can support text
         to single-view panorama or multi-view panorama generation.}
        \label{fig:framework}
    \end{center}
    \vspace{-0.6cm}
\end{figure}

\section{Proposed Method: DiffPano}
\label{diffpano}
% DiffPano可以根据相机视角和文本描述生成多视角相关联的全景图像如图1所示。首先，我们将在3.1中介绍基于全景图的单视角Stable Diffusion。然后在3.2中阐述如何利用球形对极注意力模块将单视角全景生成扩展到多视角一致的全景生成。
DiffPano is capable of generating multi-view consistent panoramas conditioned on camera viewpoints and textual descriptions, as illustrated in Fig.~\ref{fig:teaser}. In this section, we first introduce our single-view panorama stable diffusion in Sec.~\ref{pano_sd}. We then elaborate on how to extend single-view panorama generation to multi-view consistent panorama generation by leveraging the spherical epipolar attention module in Sec.~\ref{multi}.

\subsection{Single-View Panorama-Based Stable Diffusion}
\label{pano_sd}
% Stable Diffsion在训练过程中运用了大量透视图和与之对应的文本描述作为训练数据，因此具备非常好的透视图像先验知识和文本理解能力。由文本生成全景图像的任务可以视为对Stable Diffusion生成的图像的风格进行改变，由透视图转换为全景图风格。

% To generate a single-view panorama, a straightforward approach is to directly train a text-to-panorama diffusion model from scratch using plenty of text-panorama pairs, which is time-consuming and uneconomical. Stable Diffusion~\cite{stable_diffusion} leverages a vast amount of perspective images and their corresponding textual descriptions as training data, endowing it with excellent prior knowledge of perspective imagery and strong text comprehension capabilities. The task of generating panoramas from text can be regarded as a style transformation of the images generated by Stable Diffusion, converting them from a perspective style to a panoramic style. 

A straightforward way to generate a single-view panorama from text is to train a text-to-panorama diffusion model from scratch with a large number of text-panorama pairs, which is both time-consuming and computationally expensive. However, stable diffusion~\cite{stable_diffusion} leverages a vast amount of perspective images and their corresponding textual descriptions as training data, endowing it with excellent prior knowledge of perspective images and strong text understanding capabilities. An economical and effective way for panorama generation is to fine-tune the trained perspective diffusion model with a few text-panorama pairs. To this end, panorama generation from text can be regarded as a style transfer of images generated by stable diffusion, converting them from perspective style to panoramic style, and requiring them to satisfy the left-right continuity property of panoramas.

\paragraph{LoRA-based fine-tuning}
% 文本到图像的扩散模型具有非常好的二维图像先验知识和文本理解能力，我们希望保留模型的这些能力，通过微调来让生成的图像转变为全景图风格。我们采用了LoRA微调方法，这是一种之前在大语言模型中使用的微调方法，它相比全量微调速度更快，且所需的计算资源更少。我们冻结了原来Stable Diffusion中的所有参数，通过LoRA微调方法在模型的UNet中添加可训练的层，利用我们自己创建的全景图文本数据集进行训练。提高生成图像的左右连续性，我们对训练数据中的全景图进行数据增强，随机将图像的右边一部分拼接到左边。实验表明通过这种方法生成的全景图像具备不错的左右连续性。
Diffusion models for text-to-image generations possess excellent prior knowledge of 2D images and strong text comprehension capabilities. We aim to preserve these abilities of the model while fine-tuning it to generate images in the style of panoramas. We employ the Low-Rank Adaptation (LoRA)~\cite{lora} fine-tuning method, which has been previously used in large language models. Compared to full fine-tuning, LoRA is faster and requires fewer computational resources. In our approach, we freeze all the parameters of the original Stable Diffusion model and add trainable layers to the UNet component using the LoRA fine-tuning method. We then train the model using our custom-created panorama-text dataset. To improve the left-right continuity of the generated images, we perform data augmentation on the panorama training dataset by randomly concatenating a portion of the right side of the panorama to the left side. Experiments demonstrate that the panoramas generated using this method exhibit satisfactory left-right continuity.

\subsection{Spherical Epipolar-Aware Multi-View Diffusion}
\label{multi}
% 基于我们提出的single-view pano-based Stable Diffusion，我们将扩散模型改为多分支结构，并于引入Spherical Epipolar-Aware模块来生成多视角一致且可扩展的全景图。
Built upon our proposed single-view panorama stable diffusion in Sec.~\ref{pano_sd}, we extend the single-view diffusion model to a multi-view diffusion model with a spherical epipolar-aware attention module to generate multi-view scalable and consistent panoramas.

% \paragraph{Epipolar attention}
% % 对极注意力是通过在两个不同相机视角的图像之间计算对极约束实现的。对于目标视图上的每一个像素点，根据目标视图与参考视图之间的相对位姿关系计算出该像素点在参考视图中对应的极线，也就是该像素点在参考视图中可能出现的位置。这本质上是因为我们无法得知像素点实际对应的深度值，而无法计算出在参考视图中的准确位置，极线可以被视为在近平面和远平面之间根据所有假设的不同深度来采样的点的集合。对深度值进行采样等同于在在世界坐标系下的相机中心到图像像素点的射线中采样，通过将采样点反投影到参考视图中寻找对应的像素特征，再将所有采样点获取的特征进行聚合作为目标视图的特征值，从而实现目标视图与参考视图之间的对极注意力。
% Epipolar attention is achieved by computing the epipolar constraint between images from two different camera views. For each pixel in the target view, the corresponding epipolar line in the reference view is calculated based on the relative pose between the target and reference views. This line represents the possible locations where the pixel can appear in the reference view. Essentially, since we cannot determine the actual depth value corresponding to the pixel, it is impossible to compute its precise location in the reference view. The epipolar line can be considered as a collection of points sampled between the near and far planes according to all hypothesized depths.Sampling depth values is equivalent to sampling points on the ray from the camera center to the image pixel in the world coordinate system. By reprojecting the sampled points into the reference view, we search for corresponding pixel features. The features obtained from all sampled points are then aggregated as the feature value of the target view, thereby realizing the epipolar attention between the target and reference views.

\paragraph{Spherical Epipolar-Aware Attention Module}
% Epidiff提出了epipolar attention保证生成的多视角透视图之间一致性，但是由于透视图和全景图成像方式的差异，已有的epipolar attention是无法直接用于全景图的。为了解决这个挑战，我们重新推导了对于全景图的极线形式，具体的证明过程在附录里，~\ref{spherical_epipolar_line_proof}
\label{Spherical epipolar attention}
Epipolar attention was proposed in ~\cite{epidiff, pose_guided} to ensure consistency between generated multi-view perspective images. However, due to the differences in imaging methods between perspective and panoramic views, existing epipolar attention cannot be directly used for panoramic views. To overcome this challenge, we derived the epipolar line for panoramic images in the equirectangular projection (ERP), and the specific proof process is provided in Appendix~\ref{sec:spherical_epipolar_line_proof}. Equation~\ref{eq:epipolar} shows the mathematical form of the spherical epipolar line in ERP images. The spherical epipolar line is visualized in the spherical epipolar-aware attention module of Fig.~\ref{fig:framework}.
% 我们通过将对极注意力的原理应用到全景图中实现Spherical epipolar-aware module。全景图和透视图一个非常大的区别就是它是采用球形投影的方式成像而非平面成像。给定一个目标视图中的像素点p，根据球形投影的过程求出它对应的球面坐标p_sphere：theta = (0.5 - x_pix) * 2 * π，phi = -((y_pix - 0.5) * π)，其中px,py为p的像素坐标，theta和phi为其对应的球面坐标。再将球面坐标系转化为笛卡尔坐标系来得到p对应的相机坐标p_camera：x_cam = cos(phi) * sin(theta)，y_cam = sin(phi)，z_cam = cos(phi) * cos(theta)。将p的相机坐标通过相机的位姿矩阵转换为世界坐标p_world，从而计算出在世界坐标系下，相机中心到p_world的射线，并通过3.2.1中的方法构造对极注意力。基于全景图的极线计算方法将在补充材料中介绍。
We extend the principle of epipolar attention to panoramic images to implement the spherical epipolar-aware attention module. Given a pixel $\mathbf{p}$ in the target view, we calculate its corresponding spherical coordinates $\mathbf{p}_{sphere}$ based on the spherical projection process:

% One significant difference between panoramic and perspective images is that the former uses spherical projection for imaging instead of planar projection.

\begin{equation}
\label{sphere_cord}
\begin{aligned}
\theta &= (0.5 - \frac{x_{pix}}{W}) \cdot 2\pi \\
\phi &=(0.5 - \frac{y_{pix}}{H}) \cdot \pi,
\end{aligned}
\end{equation}
where $x_{pix}$ and $y_{pix}$ are the pixel coordinates of $\mathbf{p}$,  $\theta$ and $\phi$ are its corresponding spherical coordinates and $W$ and $H$ are the resolutions of panorama. We then transform the spherical coordinate system to the Cartesian coordinate system to obtain the camera coordinates $p_{camera}$ corresponding to $\mathbf{p}$ :

\begin{equation}
\label{cam_cord}
\begin{aligned}
x_{cam} &= \cos(\phi) \cdot \sin(\theta) \\
y_{cam} &= \sin(\phi) \\
z_{cam} &= \cos(\phi) \cdot \cos(\theta).
\end{aligned}
\end{equation}

The camera coordinates of $\mathbf{p}_{cam}$ are converted to world coordinates $\mathbf{p}_{world}$ through the camera's pose matrix. This allows us to compute the ray from the camera center to $\mathbf{p}_{world}$ in the world coordinate system and construct the spherical epipolar attention module.
% using the method described in Section~\ref{multi}. 
% The method for computing spherical epipolar lines based on panoramas will be introduced in the supplementary material.

% 给定N张全景图对应的特征图F={Fi|1<=i<=N},及其对应的相机位姿矩阵，我们通过Spherical epipolar-aware module实现不同视角之间的交叉注意力。在生成过程中，F中的每一个都可以被视为目标视角，并在其余特征中选择最邻近的K个视角作为参考视角。对于目标视角特征图中的每一个特征点，在它与相机之间的光线上均匀采样S个点，并将所有采样点反投影到所有K个参考视角的特征图中，通过差值获取相应特征值。将目标视图中的特征视为q，获取的所有采样点在参考视图中的特征作为k和v，构造交叉注意力。
Given $N$ feature maps $F={F_i|1\leq i\leq N}$ corresponding to $N$ panoramic images and their respective camera pose matrices, we implement cross-attention between different views through the spherical epipolar-aware module. During the generation process, each feature map in $F$ can be considered as the target view, and the $K$ nearest views are selected from the remaining features as reference views.

For each feature point in the target view feature map, we uniformly sample $S$ points on the ray between the feature point and the camera. All sampled points are reprojected onto the feature maps of the $K$ reference views, and the corresponding feature values are obtained through interpolation. We denote the features in the target view as $q$, and the features of all sampled points in the reference views as $k$ and $v$. The cross-attention is then constructed using these features.

Let $p_i$ be a feature point in the target view feature map $F_t$, and ${p_{i,j}|1\leq j\leq S}$ be the $S$ uniformly sampled points on the ray between $p_i$ and the camera center. We reproject these points onto the $K$ reference view feature maps ${F_{r_k}|1\leq k\leq K}$ to obtain the corresponding feature values ${f_{i,j,k}|1\leq j\leq S, 1\leq k\leq K}$. The query $q_i$, key $k_i$, and value $v_i$ for the cross-attention mechanism are defined as follows:

\begin{equation}
q_i = F_t(p_i), \quad k_i = {f_{i,j,k}|1\leq j\leq S, 1\leq k\leq K}, \quad v_i = {f_{i,j,k}|1\leq j\leq S, 1\leq k\leq K}.
\end{equation}

The cross-attention output $o_i$ for the feature point $p_i$ is computed as:
\begin{equation}
o_i = Attention(q_i, k_i, v_i) = softmax(\frac{q_i k_i^T}{\sqrt{d}})v_i,
\end{equation}
where $d$ is the dimension of the query and key vectors.

\paragraph{Positional Encoding}
% 参考epi中Positional encoding，只使用了absolute positioning。
% 为了增强模型对于不同视角之间的3D空间理解能力，我们运用了 Light Field Networks (LFN)中的位置编码方法。在世界坐标系下，每一个像素点与相机之间的光线方向向量被转换为Pl¨ucker坐标r = (o × d, d)，该坐标由光线对应的起点o和方向d组成。Pl¨ucker坐标表示的光线和采样点对应的深度分别通过harmonic transformation进行转换再将两者结合，并与特征图进行结合，增强其中的3D空间信息，便于模型学习不同视角之间的空间位置关系。
To enhance the model's understanding of 3D spatial relationships between different views, we follow EpiDiff~\cite{epidiff} to employ the positional encoding method from Light Field Networks (LFN)~\cite{LFN}. In the world coordinate system, let $p_i$ be a pixel in the target view, and $o_i$ and $d_i$ be the origin and direction of the ray between $p_i$ and the camera center, respectively. The Plücker coordinates $r_i$ of the ray are computed as:
\begin{equation}
r_i = (o_i \times d_i, d_i).
\end{equation}

% the direction vector of the ray between each pixel $p_i$ and the camera center is converted to Plücker coordinates $r_i = (o_i \times d_i, d_i)$, which consist of the origin $o_i$ and direction $d_i$ corresponding to the ray. The Plücker coordinates representing the ray and the spherical depth corresponding to the sampled points are separately transformed using a harmonic transformation, and then combined. The resulting features are concatenated with the feature maps to enhance the 3D spatial information, facilitating the model's learning of spatial relationships between different views. The major difference is that we emit the ray according to spherical projection instead of perspective projection. 
% For each sample point, we embed the code of the spherical depth instead of the planar depth.
For each sampled point $p_{i,j}$ on the ray, its corresponding spherical depth $z_{i,j}$ is transformed using a harmonic transformation to get $\gamma_z(z_{i,j})$. Similarly, the Plücker coordinates $r_i$ are transformed as $\gamma_r(r_i)$.The positionally encoded features $\gamma_r(r_i)$ and $\gamma_z(z_{i,j})$ are then concatenated to obtain the combined positional encoding $\gamma(r_i, z_{i,j})$:
\begin{equation}
\gamma(r_i, z_{i,j}) = [\gamma_r(r_i), \gamma_z(z_{i,j})].
\end{equation}
The combined positional encoding $\gamma(r_i, z_{i,j})$ is then concatenated with the feature maps $F_t$ and ${F_{r_k}|1\leq k\leq K}$ to obtain the enhanced feature representations $\hat{F}t$ and ${\hat{F}{r_k}|1\leq k\leq K}$:
\begin{equation}
\hat{F}t(p_i) = [F_t(p_i), \gamma(r_i, z_{i,j})], \quad \hat{F}{r_k}(p_{i,j}) = [F_{r_k}(p_{i,j}), \gamma(r_i, z_{i,j})],
\end{equation}

where $[\cdot, \cdot]$ denotes concatenation. These enhanced feature representations are then used in the cross-attention mechanism to improve the model's understanding of 3D spatial relationships between different views.

\paragraph{Two-Stage Training}
% 全景图与透视图之间最大的区别就是全景图包含了周围360°的内容，而透视图只有给定视角下的内容，因此，当相机只进行旋转或者平移量较小时，对应全景图中的内容几乎没有什么变化。为了让生成的多视角全景图像与其对应的每一个文本更相匹配，我们将训练划分为两个阶段。在第一阶段中我们使用图像内容几乎没有变化（相机移动的幅度较小）的数据进行训练，在这个过程中提升spherical epipolar-aware module的效果。在第二阶段中增加每个视角之间的相机移动距离，用产生新内容的图像进行训练，提升模型在感觉空间位置变化的基础上对文本的理解能力，在保证多视角一致性的同时，增强可扩展性。
The main difference between panoramic images and perspective images is that panoramic images contain $360^{\circ}$ content of the surroundings, while perspective images only contain content from a given viewpoint. Therefore, when the camera only rotates or translates by a small amount, the corresponding content in the panoramic image hardly changes. To make the generated multi-view panoramic images better match each corresponding text, we divide the training into two stages. In the first stage, we use the selected dataset with almost no change in image content (small camera movement) for training, which enhances the effect of the spherical epipolar-aware attention module. In the second stage, we increase the camera movement distance between each viewpoint and train with images that generate new content, improving the model's ability to understand text based on changes in perceived spatial location while ensuring multi-view consistency and enhancing scalability.

\section{Experiment}
\paragraph{Dataset}
% 我们通过Habitat Simulator在包含900个室内场景的有史以来最大的3D空间数据集Habitat Matterport 3D（HM3D）数据集中渲染数据。在基于全景图的Stable Diffusion任务中，我们选取了HM3D中的100个场景，在场景中以随机视角渲染立方体贴图，并通过差值拼接成全景图。我们过滤了生成全景图中因原始网格问题导致黑色像素占比较多的图像，最终生成了包含8508张全景图的数据集。在pano-based multi-view synthesis任务中，我们在完整900个场景中根据相机轨迹渲染图像。为了适应第一阶段训练，我们选取了变化较小的相机移动轨迹。而对于第二阶段训练的数据，我们在一条较长的相机轨迹中比较每一个视角渲染得到的全景图.每一组训练数据由4帧全景图像和其对应的位姿矩阵以及文本描述组成。经过筛选，图像内容几乎相同的数据中包含19739组数据，而图像内容不相同的数据中有18704组数据。
We leverage the Habitat Simulator~\cite{habitat} to render a panoramic video dataset based on the Habitat Matterport 3D (HM3D) dataset~\cite{hm3d}. The pipeline of dataset rendering and captioning is shown in Sec.~\ref{Pano Video-Text Dataset}. After post-processing such as dataset filtering, we constructed 8,508 panorama-text pairs as training sets for single-view panorama generation. For multi-view panorama generation, we constructed 19,739 multi-view panorama-text pairs with nearly identical image content and 18,704 multi-view panorama-text pairs with different image contents as training sets. For specific details regarding the dataset, please refer to Appendix~\ref{dataset_process}. 

\paragraph{Implementation Details}
% 在pano-based multi-view synthesis中，我们一次性生成N=4张不同视角的全景图，在Spherical epipolar-aware module中将与目标视图邻近的三个视图作为参考视图，即K=3，同时在每条光线上采样S=10个点。分别对图像内容几乎相同和图像内容不同的数据各训练100轮。
In the multi-view panorama generation, we simultaneously generate $N=4$ panoramas from different viewpoints. Within the spherical epipolar-aware attention module, we consider the two nearest views to the target view as reference views, i.e., $K=3$, and sample $S=10$ points along each ray. We conducted separate training for 100 epochs on datasets with almost identical image content and datasets with varying image content. 
Please refer to Appendix~\ref{network_detail} for further implementation details.

\paragraph{Evaluation Metrics}
% 对于我们提出的single-view pano-based Stable Diffusion，我们利用FID、IS和CS指标进行评价，其中FID和IS用于评测生成全景图的真实性和多样性，CS用来评测文本与图像之间的一致性。
To evaluate the performance of our proposed single-view panorama-based stable diffusion model, we employ three commonly used metrics: Fréchet Inception Distance (FID)~\cite{FID}, Inception Score (IS)~\cite{IS}, and CLIP Score (CS)~\cite{CS}. FID measures the similarity between the distribution of generated panoramas and the distribution of real images. IS assesses the quality and diversity of the generated panoramas. CS is utilized to evaluate the consistency between the input text and the generated panoramas. To further evaluate the consistency of multi-view panorama generation, we leverage the Peak Signal-to-Noise Ratio (PSNR) and Structural Similarity Index Measure (SSIM)~\cite{SSIM} metrics. PSNR and SSIM quantify the pixel-level differences and structural similarity between the generated views respectively.

\subsection{Single-View Panorama Generation}

\begin{figure}[ht]
    \begin{center}
        \centerline{\includegraphics[width=1\linewidth]{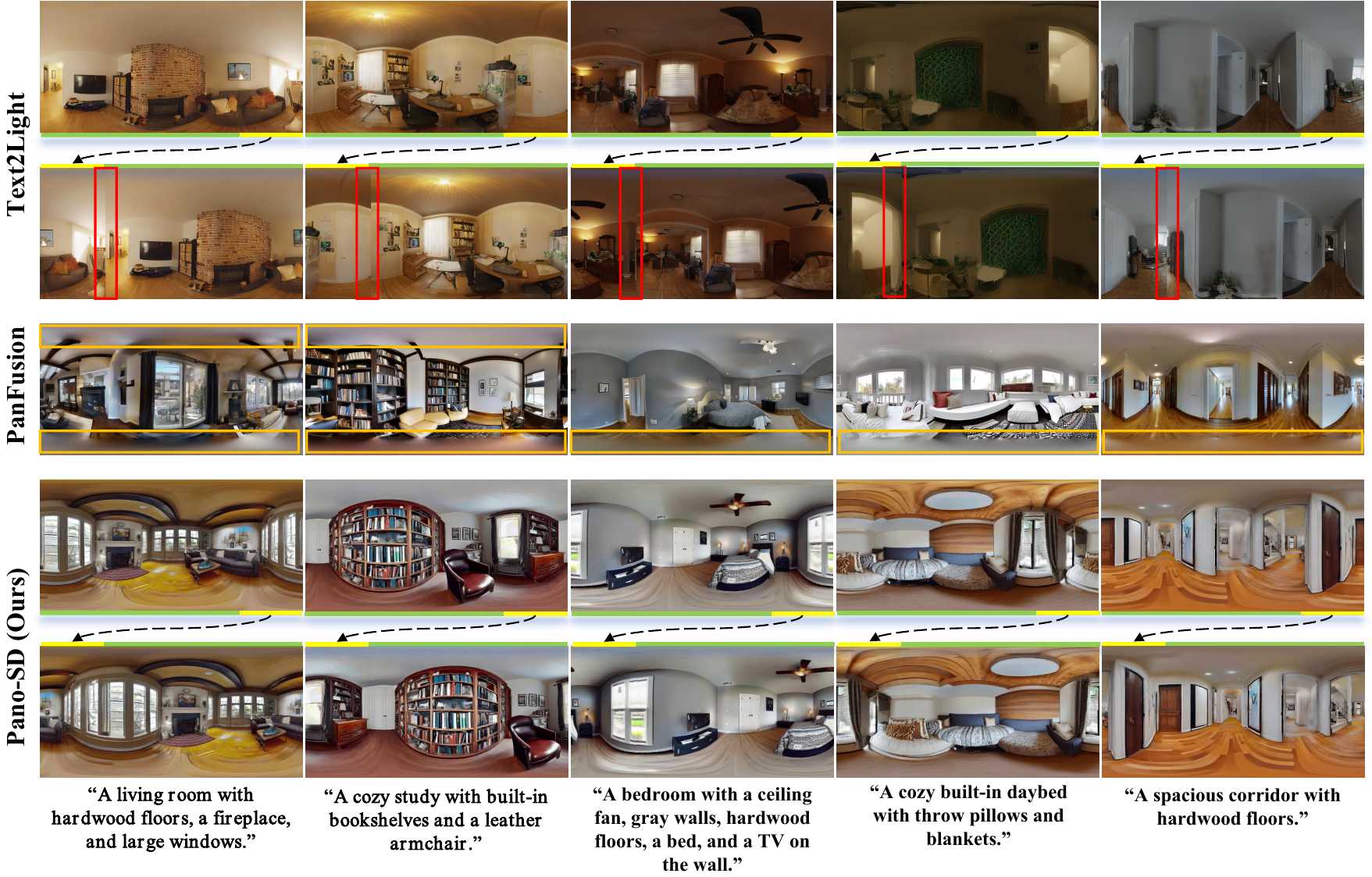}}
        \vspace{-0.1cm}
        \caption{\textbf{Text to Panorama Comparison between TextLight, PanFusion, and Ours.} }
        \label{fig:single_pano}
    \end{center}
    \vspace{-0.6cm}
\end{figure}

\paragraph{Baselines}
%我们与下面这些baseline方法比较全景图像生成的效果：
We evaluate the performance of our proposed method by comparing it with the following baseline approaches for text-to-panorama generation:

% Text2Light:先基于文本生成低分辨率的全景图，再扩展到超高分辨率。PanFusion：双分支文本到全景图模型, 减轻了透视图投影到全景图中的失真的同时提供全局布局指导。
\begin{itemize}
\item $\mathit{Text2Light}$~\cite{text2light}is a two-stage approach that first generates a low-resolution panorama based on the input text, and then expands it to ultra-high resolution. 
\item $\mathit{PanFusion}$~\cite{panfusion}is a dual-branch text-to-panorama model that aims to mitigate the distortion caused by projecting perspective images onto a panoramic canvas while providing global layout guidance. 
\end{itemize}
Since the panoramic images generated by MVDiffusion~\cite{mvdiffusion} do not include top and bottom viewpoints, they are not complete panoramas. Therefore, we do not compare our method with MVDiffusion.

\paragraph{Quantitative Results}
% 表1展示的是定量的结果。Text2Light的FID、IS和CS值是引用PanFusion中的结果。我们使用PanFusion中的数据集训练了我们提出的Pano-SD,并重新测试了PanFusion的效果。从结果中可以看出，我们的方法在CS值上略有领先，FID值远胜于Text2Light,与PanFsion接近。由于我们的模型结构比较简单且数据集图像中的底部和顶部模糊会影响生成的整体图像质量，所以在IS值上与PanFusion有一定差距。然而我们的方法在推理时间上有巨大的优势，这可以大大提高后续多视角生成任务的效率。
Table~\ref{pano_sd Quantitative} presents the quantitative results. The FID, IS, and CS values of Text2Light are from the PanFusion~\cite{panfusion}. We trained our proposed Pano-SD using the same dataset as PanFusion~\cite{panfusion} and re-evaluated the performance of PanFusion. From the results, it can be observed that our method slightly outperforms others in terms of CS value and achieves a significantly lower FID value compared to Text2Light, which is close to PanFusion~\cite{panfusion}. Moreover, due to the simplicity and efficiency of our model architecture, our method has a substantial advantage in inference time, which can significantly improve the efficiency of subsequent multi-view generation tasks.

\paragraph{Qualitative Results}
% 定性比较的结果如图所示。我们比较了三个模型在各自数据集上训练后的结果。Text2Light生成的全景图在左右一致性上的效果比较差。而PanFusion生成的全景图的底部和顶部会存在模糊的情况，影响了全景图像的整体性。而我们的模型能够生成底部和顶部清晰的且左右连续性较好的全景图，但由于数据集质量的缘故，在图像质量上会稍差一些。
The qualitative comparison results are shown in Figure~\ref{fig:single_pano}. We compare our method with the two models trained on their respective datasets. The panoramas generated by Text2Light exhibit poor left-right consistency. On the other hand, the panoramas generated by PanFusion suffer from blurriness at the bottom and top regions, which affects the overall integrity of the panoramas. In contrast, our model is capable of generating panoramas with clear bottom and top regions and better left-right continuity. However, due to the quality of the dataset, the image quality may be slightly inferior.

\begin{table}
  \caption{Quantitative Panorama Comparisons with Baseline Methods}
  \label{pano_sd Quantitative}
  \centering
  \begin{tabular}{ccccc}
    \toprule
    Method  & FID$\downarrow$  & IS$\uparrow$  & CS$\uparrow$  & Inference time(s)$\downarrow$\\
    \midrule
    Text2Light &76.50 &3.60 &27.48 &50.4     \\
    PanFusion  &\bf{47.62} &\bf{4.49} &28.73 &27.6 \\
    Pano-SD(Ours) &48.52 &3.30 &\bf{28.98} &\bf{5.1}\\
    \bottomrule
  \end{tabular}
\end{table}

% MVDiffusion &188.16	&1.82 \\
%     PanFusion  &213.19 &\bf{2.61}  \\
%     Pano-SD(Ours) &\bf{164.75} &1.96\\

\subsection{Multi-View Panorama Generation}

% 我们去除Spherical epipolar-aware module，载入pano_sd中预训练的LoRA层，将unet中的self-attention转换为MVDream中的形式。同时，将相机位姿矩阵通过2层MLP转为camera embeddings并作为残差加入time embeddings。实验的定性结果如图所示，在相同的训练批数下，DiffPano在多视角一致性上的效果远胜于MVDream中的方法，我们的方法能够在多视角图像的细节上保持一致性，而MVDream中的方法只能在图像整体上具有一定相似性。即使与两倍训练轮次的结果相比，我们的方法在一致性上仍然效果更好。

To the best of our knowledge, there is no method for multi-view panorama generation, the existing SOTA method MVDream for perspective images cannot be directly applied to multi-view panorama generation. To verify the validity of our proposed spherical epipolar-aware attention module, we adapted MVDream to the panorama generation task as a comparative baseline.

\begin{figure}[ht]
    \begin{center}
        \centerline{\includegraphics[width=1\linewidth]{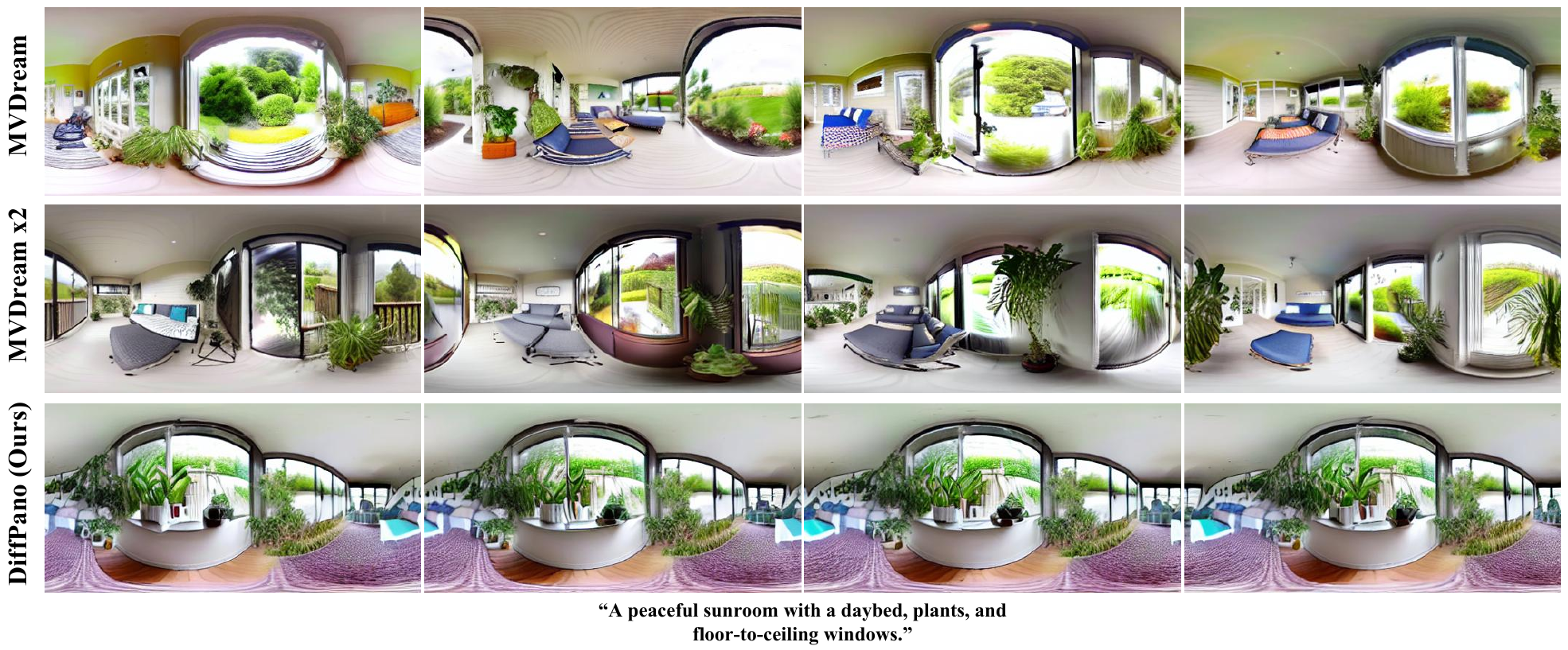}}
        \vspace{-0.1cm}
        \caption{\textbf{Comparisons with MVDream.} DiffPano can generate more consistent multi-view panoramas. "MVDream$\times$2" denotes MVDream is trained with twice iteration number relative to our method.}
        \label{fig:cmp_mvdream}
    \end{center}
    \vspace{-0.6cm}
\end{figure}

Specifically, we remove the spherical epipolar-aware attention module from our method and load the pre-trained LoRA layers from Pano-SD. We then convert the 3D self-attention in the UNet to the form used in MVDream~\cite{mvdream}. Additionally, we transform the camera pose matrix into camera embeddings through a 2-layer MLP and add it as a residual to the time embeddings. The qualitative comparison of the experiment is shown in Fig~\ref{fig:cmp_mvdream}. Under the same training iteration, DiffPano significantly outperforms MVDream~\cite{mvdream} in terms of multi-view consistency. Our method can maintain consistency in the details of multi-view images, while MVDream can only achieve a certain level of similarity in the overall images. Even compared to MVDream with twice the number of training iterations, our method still performs better in terms of consistency.

\paragraph{User Study}
We collected 20 text prompts and recruited nearly 50 volunteers to evaluate text-to-multi-view panoramic image generation. Evaluation metrics of multi-view ERP panorama generation include the quality of multi-view panoramic images, multi-view consistency, and the consistency between the text and multi-view panoramas. Experimental results on Tab.~\ref{tab:user_study} show that our method can generate multi-view panoramic images with better quality, higher text and image similarity, and more consistent multi-view images, compared with MVDream~\cite{mvdream} and PanFusion~\cite{panfusion}. See more qualitative results in Fig.~\ref{fig:ablation study} to Fig.~\ref{fig:ablation study5}. 

\subsection{Ablation Study}

\begin{table}
  \caption{User Study of Text to Multi-view Panoramas}
  \label{tab:user_study}
  \centering
  \begin{tabular}{cccc}
    \toprule
    Method  & Image quality$\uparrow$  
    & Image-text consistency$\uparrow$  
    & Multi-view consistency$\uparrow$  \\
    \midrule
    MVDream &3.6 & 3.7 & 3.8     \\
    \midrule
    PanFusion  & 3.8 & 4.0 & 3.0  \\
    \midrule
    DiffPano(Ours) &\textbf{4.0} & \textbf{4.2} & \textbf{4.3}\\
    \bottomrule
  \end{tabular}
  
\end{table}

\begin{table}
  \caption{Ablation Study on the Number of Sampling Points and Reference Frames}
  \label{num_ref_sample Quantitative}
  \centering
  \begin{tabular}{lcccccc}
    \toprule
      & FID$\downarrow$  & IS$\uparrow$  & CS$\uparrow$  &PSNR$\uparrow$ &SSIM$\uparrow$ &Inference time(s)$\downarrow$ \\
    \midrule
    $\mathit{K}$ = 1, $\mathit{S}$ = 10 
    &73.30 &3.40 &22.14 &29.99 &0.69 &30.06      \\
    $\mathit{K}$ = 2, $\mathit{S}$ = 10
    &66.02 &3.34 &22.92 &32.04 &0.79 &33.12        \\
    $\mathit{K}$ = 3, $\mathit{S}$ = 10
    &69.89 &3.58 &22.76 &32.29 &0.81 &35.79       \\
    \midrule
    $\mathit{K}$ = 4, $\mathit{S}$ = 6
    &68.39 &3.57 &22.74 &33.32 &0.86 &26.61        \\
    $\mathit{K}$ = 4, $\mathit{S}$ = 8
    &67.30 &3.54 &22.66 &33.00 &0.84 &32.23        \\
    $\mathit{K}$ = 4, $\mathit{S}$ = 10
    &65.98 &3.37 &22.59 &33.39 &0.87 &37.91        \\
    $\mathit{K}$ = 4, $\mathit{S}$ = 12
    &62.79 &3.26 &22.65 &32.89 &0.83 &43.72       \\
    \bottomrule
  \end{tabular}
\end{table}

\paragraph{Number of Reference Views and Sample Points}
% 1帧 2帧 3帧 4帧(50epoch) 256*512
% 不同的参考帧数量意味着模型可以从不同数量的特征中聚合信息。在相机到图像的光线上采样点的数量对模型的效果也有一定的影响。我们一次性生成5个视角的全景图，计算每组第一帧图像的FID、IS、CS指标来衡量模型在不同数量的参考帧和采样点下生成的图像的质量。并将每组生成的全景图的第一帧和最后一帧设置相同的相机位姿，通过计算这两帧图像之间的PSNR和SSIM值来来衡量模型的一致性。
% 结果如表所示，提高参考帧的数量和采样点数可以在一定程度上提高生成全景图的质量，但在图像多样性和与文本的一致性上变化不大。随着参考帧数和采样点数的增加，模型的一致性有所提高，但当采样点数过多(S=12)时会降模型的多视角一致性.
To explore the influence of varying numbers of reference frames and sampling points on model performance, we generate 5-frame multi-view panoramas simultaneously, and compute the FID, IS, and CS metrics for the first frame of each group to assess the quality of the generated panorama under different quantities of reference frames and sampling points. Furthermore, we set the same camera pose for the first and last frames of each generated panorama group, and calculate the PSNR and SSIM values between these two frames to evaluate the model's multi-view consistency.
As shown in Table.~\ref{num_ref_sample Quantitative}, increasing the number of reference frames and sampling points can improve the quality of generated panoramas to a certain extent, but the changes in image diversity and consistency with text remain marginal. With an increasing number of reference frames and sampling points, the model's consistency exhibits improvement, however, when the number of sampling points becomes excessive ($\mathit{S}$=12), the multi-view consistency of the model diminishes.

\begin{table}
  \caption{Ablation Study of One-Stage vs Two-Stage Training}
  \label{One-Stage vs Two-Stage Training}
  \centering
  \begin{tabular}{lccccc}
    \toprule
      & FID$\downarrow$  & IS$\uparrow$  & LPIPS$\downarrow$  &PSNR$\uparrow$ &SSIM$\uparrow$ \\
    \midrule
    One-stage & 82.92 &4.52		&0.0454	&31.64	&0.74 \\
Two-stage &74.08 &	3.13	&	0.0610	& 31.54	&0.73 \\
    \bottomrule
  \end{tabular}
\end{table}

\paragraph{One-Stage vs Two-Stage} We conduct ablation experiments on one-stage and two-stage training. Table.~\ref{One-Stage vs Two-Stage Training} shows that the two-stage method will obtain the higher FID values.
% due to the introduction of the epi module, the multi-view consistency of the two training methods is relatively good. Comparing the consistency is not enough to measure the difference between the two methods. 
The IS of the two-stage training method is lower, and the diversity is reduced to a certain extent, which is slightly worse than the one-stage training. However, it should be noted that the images after one-stage training will have ghosting, but the two-stage will not. 
% We show the qualitative comparison results in the global response (the FID value of the two-stage method will also be higher).

\section{Conclusion}
We have proposed the panoramic video-text dataset and panorama generation framework with spherical epipolar-aware attention for text-to-single-view or multi-view panorama generation. Extensive experiments demonstrate that our method can achieve scalable, consistent, and diverse multi-view panoramas.
\paragraph{Limitation and Future Work}
% 多帧当推理出的帧数超过训练时出的帧数时，会逐渐凭空出现新内容，一致性也会下降。
Although our method demonstrates the ability to generate consistent multi-view panoramas under the same setting as the training phase, it is important to note that as the number of frames increases during inference, the model tends to hallucinate content. 

Exploring the use of video diffusion models to improve the consistency of generated multi-view panoramas is a promising direction. Longer panoramic videos are expected to be realized based on the generated panoramas as conditions.

% \paragraph{Future works}
% 探索能否通过图像作为条件的形式进一步提高一致性，并增加生成帧数。
% Building upon the foundation of our current research, we intend to pursue further avenues of exploration to address the challenges associated with maintaining consistency across generated multi-view panoramas and achieving more remarkable outcomes. One promising direction lies in examining the potential benefits of incorporating images as a form of conditioning within the generative process.

\section{Acknowledgements}
This work was partially supported by the National Key Research and Development Program of China (No. 2023YFF0905104) and National Natural Science Foundation of China (Nos. 61932003, 62076184 and 62473286).

\bibliographystyle{plain}
\bibliography{neurips_2024}

\newpage

\appendix

\section{Spherical Epipolar Line Computation}
\label{sec:spherical_epipolar_line_proof}
% 给定目标视角i下的全景图中一点p的像素坐标，可以计算出该点在源视角j下的全景图中对应的极线。首先根据公式(2)(3)，计算出p点在目标视角下对应的相机坐标p_cam = (x_cam, y_cam, z_cam)^T。根据相对位姿{R^i→j,T^i→j},计算出p点在源视角j下的相机坐标p'_cam：p'_cam = R^i→jp_cam+T^i→j,同时计算出目标视角i下的相机原点o=(0,0,0)^T投影到源视角j下的坐标o':o'=R^i→jo+T^i→j.
% 我们需要找到在源视角j的相机坐标系下p',o,o'三点所在的平面L，该平面与源视角下的球面的交线即为所求的极线。平面L对应的方程为：AX+BY+CZ+D=0，代入p',o,o'三点的坐标可得：
Given the pixel coordinates of a point in the panorama of the target view $i$, the corresponding epipolar line in the panorama of the source view $j$ can be calculated. First, according to Eq. (~\ref{sphere_cord}), the camera coordinates $\mathbf{p}$ in the target view $i$ are computed. Based on the relative pose $\left \{ {\mathbf{R}^{i \rightarrow j}, \mathbf{T}^{i \rightarrow j}} \right \} $, $\mathbf{p'}$ in the source view $j$ are calculated:

\begin{equation}
\mathbf{p'} = \mathbf{R}^{i \rightarrow j}\mathbf{p} + \mathbf{T}^{i \rightarrow j}.
\end{equation}

Simultaneously, the coordinates $\mathbf{o'}$ of the camera origin $\mathbf{o} = (0, 0, 0)^T$ in the target view $i$ projected onto the source view $j$ are computed:

\begin{equation}
\mathbf{o'} = \mathbf{R}^{i \rightarrow j}\mathbf{o} + \mathbf{T}^{i \rightarrow j}.
\end{equation}

Then, we need to find the plane $L$ containing the three points $\mathbf{p'}$, $\mathbf{o}$, and $\mathbf{o'}$ in the camera coordinate system of the source view $j$. The intersection of this plane with the spherical surface is the desired epipolar line. The equation corresponding to plane $L$ is:

\begin{equation}
AX + BY + CZ + D = 0.
\end{equation}

Substituting the coordinates of the three points into the equation yields the coefficients $A$, $B$, $C$, and $D$:

\begin{equation}
\begin{aligned}
A&=\frac{z_{\mathbf{o'}} \cdot y_{\mathbf{p'}}-z_{\mathbf{p'}} \cdot y_{\mathbf{o'}}}
{x_{\mathbf{p'}} \cdot y_{\mathbf{o'}}-x_{\mathbf{o'}} \cdot y_{\mathbf{p'}}} \cdot C \\
B&=\frac{z_{\mathbf{o'}} \cdot x_{\mathbf{p'}}-z_{\mathbf{p'}} \cdot x_{\mathbf{o'}}}
{y_{\mathbf{p'}} \cdot x_{\mathbf{o'}}-
y_{\mathbf{o'}} \cdot x_{\mathbf{p'}}} \cdot C \\
D&= 0.
\end{aligned}
\end{equation}

% 为了简化表示平面L的方程，我们引入新的系数a1和a2:
To simplify the representation of the equation for plane $L$, we introduce new coefficients $a_1$ and $a_2$:

\begin{equation}
\begin{aligned}
a_1&=\frac{z_{\mathbf{o'}} \cdot y_{\mathbf{p'}}-z_{\mathbf{p'}} \cdot y_{\mathbf{o'}}}
{x_{\mathbf{p'}} \cdot y_{\mathbf{o'}}-x_{\mathbf{o'}} \cdot y_{\mathbf{p'}}} \\
a_2&=\frac{z_{\mathbf{o'}} \cdot x_{\mathbf{p'}}-z_{\mathbf{p'}} \cdot x_{\mathbf{o'}}}
{y_{\mathbf{p'}} \cdot x_{\mathbf{o'}}-
y_{\mathbf{o'}} \cdot x_{\mathbf{p'}}},
\end{aligned}
\end{equation}

the equation of plane $L$ can be simplified to:

\begin{equation}
a_1X + a_2Y + Z = 0.
\end{equation}

% 根据公式(2)(3)可得源视角j中的极线方程:其中x_pix和y_pix是对应的像素坐标。
According to Eq.\ref{sphere_cord} and Eq.\ref{cam_cord}, the epipolar line equation in the source view $j$ can be obtained:

\begin{equation}
y_{pix} = H \cdot \left[ {\arctan \left( \frac{a_1 \sin \left( \frac{2 \pi x_{pix}}{W} \right) - \cos \left( \frac{2 \pi x_{pix}}{W} \right)}{a_2} \right)} / {\pi} + 0.5 \right],
\label{eq:epipolar}
\end{equation}

where $x_{pix}$ and $y_{pix}$ are the corresponding pixel coordinates.

\section{Experiment Details}
\paragraph{Dataset Processing}
\label{dataset_process}
% 在单视角全景图像生成中，我们通过在不同视角下渲染HM3D数据集中的场景来获取全景图，但由于场景对应的mesh的质量不是特别高，存在部分缺失的情况，所以我们根据渲染图像对应的深度图进行过滤，去掉其中深度为0的部分占比较高的图像。
In single-view panorama generation, we select 100 scenes from HM3D~\cite{hm3d} and render cube maps from random viewpoints within each scene, which are then stitched into panoramas through interpolation. However, due to the imperfect quality of the corresponding scene meshes, which may have missing parts, we filter out images with a high proportion of zero-depth values based on their corresponding depth maps, ultimately creating a dataset of 8,508 panoramas.

% 为了构造图像内容几乎相同的多视角全景图数据集，我们选取了每条相机轨迹的前几帧，因为这些帧之间的相机移动幅度很小，且是在同一个场景中。图像内容发生变化的数据集需要在整条相机轨迹上构建。
In the multi-view panorama generation task, we render images based on camera trajectories across scenes. To accommodate the first stage of training, we select camera trajectories with minimal changes. In contrast, for the second stage, the dataset comprises panoramas with more significant camera displacements between consecutive frames. Each training dataset consists of multiple panoramas, their corresponding pose matrices, and text descriptions. To construct a multi-view panorama dataset with nearly identical image content, we select the first few frames from each camera trajectory, as the camera movement between these frames is minimal and they are captured within the same scene. For datasets where the image content changes, the construction needs to be performed across the entire camera trajectory. By projecting the panorama from the subsequent frame onto the viewpoint of the previous frame using spherical projection, we compare the pixel value differences between the two. If more than 40\% of the pixel values differ, we consider new content to have been generated between the two frames. After filtering, there are 19,739 data sets with nearly identical image content and 18,704 data sets with differing image content.

\paragraph{Compare with Baseline Methods}
\begin{itemize}
% 在定量分析中，FID、IS和CS数值是引用的PanFusion中的结果，是使用相同的数据集训练后测试的结果。在定性分析中我们直接使用其在室内和室外数据集上共同训练的模型来根据文本生成ldr形式的全景图像。
    \item $\mathit{Text2Light}$ \cite{text2light}: In the quantitative analysis, the FID, IS, and CS values are referenced from the results reported in PanFusion\cite{panfusion}, which were obtained by testing on the same dataset after training. For the qualitative analysis, we directly employ their jointly trained model on both indoor and outdoor datasets to generate LDR (low dynamic range) panoramas based on text descriptions.
% 为了比较生成的全景图的质量，我们也使用通过MP3D skybox图像拼接成的全景图进行训练，并使用与其相同的数据集划分和文本描述，包括9820张全景图用来训练和1092张用来评估。在定性分析中，我们使用的是自己根据HM3D构造的数据集，其中包含的全景图不存在底部和顶部模糊的情况，但是图像质量会差一些。
    \item $\mathit{PanFusion}$ \cite{panfusion}: To compare the quality of the generated panoramas, we also train the model using panoramas stitched from MP3D skybox images. We employ the same dataset split and text descriptions as in \cite{panfusion}, which includes 9,820 panoramas for training and 1,092 for evaluation. In the qualitative analysis, we use our own dataset constructed from HM3D\cite{hm3d}. The panoramas in our dataset do not suffer from blurriness at the bottom and top regions, but the overall image quality may be slightly lower.
\end{itemize}

\paragraph{Compare with MVDream}
% 我们参考MVDream中的方法，将输入self-attention的B × N × H × W × C形状的张量, format it as B × NHW × C，在计算自注意力的过程中同时综合所有视角的特征来提高生成的多视角图像之间的一致性。因为MVDream是针对物体的多视角图像生成，所以其中使用的相机位姿矩阵只包含了相机的旋转信息，而没有平移信息。而我们在生成相机嵌入的时候使用的位姿矩阵是同时包含了旋转和平移信息的，这增大了模型学习的难度，这可能也是导致使用该方法效果不是特别理想的原因之一。
We adopt the method from MVDream\cite{mvdream} by formatting the $B \times N \times H \times W \times C$ tensor input to the self-attention module as $B \times NHW \times C$. This allows the model to simultaneously integrate features from all viewpoints during the self-attention computation, thereby improving the consistency among the generated multi-view panoramas. However, since MVDream is designed for multi-view image generation of objects, the camera pose matrices used in their method only contain rotation information without translation. In contrast, when generating camera embeddings, we use pose matrices that include both rotation and translation information, which increases the learning difficulty for the model. This may be one of the reasons why using this method does not yield particularly ideal results.

\begin{table}
  \caption{Quantitative Perspective Images Comparisons with Baseline Methods}
    \label{pano_sd perspective Quantitative}
    \centering
    \begin{tabular}{ccc}
    \toprule
    Method  & FID$\downarrow$  & IS$\uparrow$  \\
    \midrule
    MVDiffusion &188.16	&1.82 \\
    PanFusion  &213.19 &\bf{2.61}  \\
    Pano-SD(Ours) &\bf{164.75} &1.96\\
    \bottomrule
  \end{tabular}
\end{table}

\paragraph{Transform Panoramas to Perspective Views} We converted the generated panoramas into perspective images and conducted quantitative comparisons, shown in Tab.~\ref{pano_sd perspective Quantitative}. Experiments show that our method achieves the lowest FID, while our method is higher than MVDiffusion in IS and slightly lower than PanFusion. It should be noted that MVDiffusion directly generates perspective images and then stitches them into panoramas. It is not in the ERP format and does not have the top and bottom parts. Our panorama generation speed is faster.

\subsection{Applications}

% 放不下补充材料。

\subsubsection{Text to Single-View Panorama}

\subsubsection{Text to Multi-View Panoramas}
\begin{figure}[ht]
    \begin{center}
    \centerline{\includegraphics[width=1\linewidth]{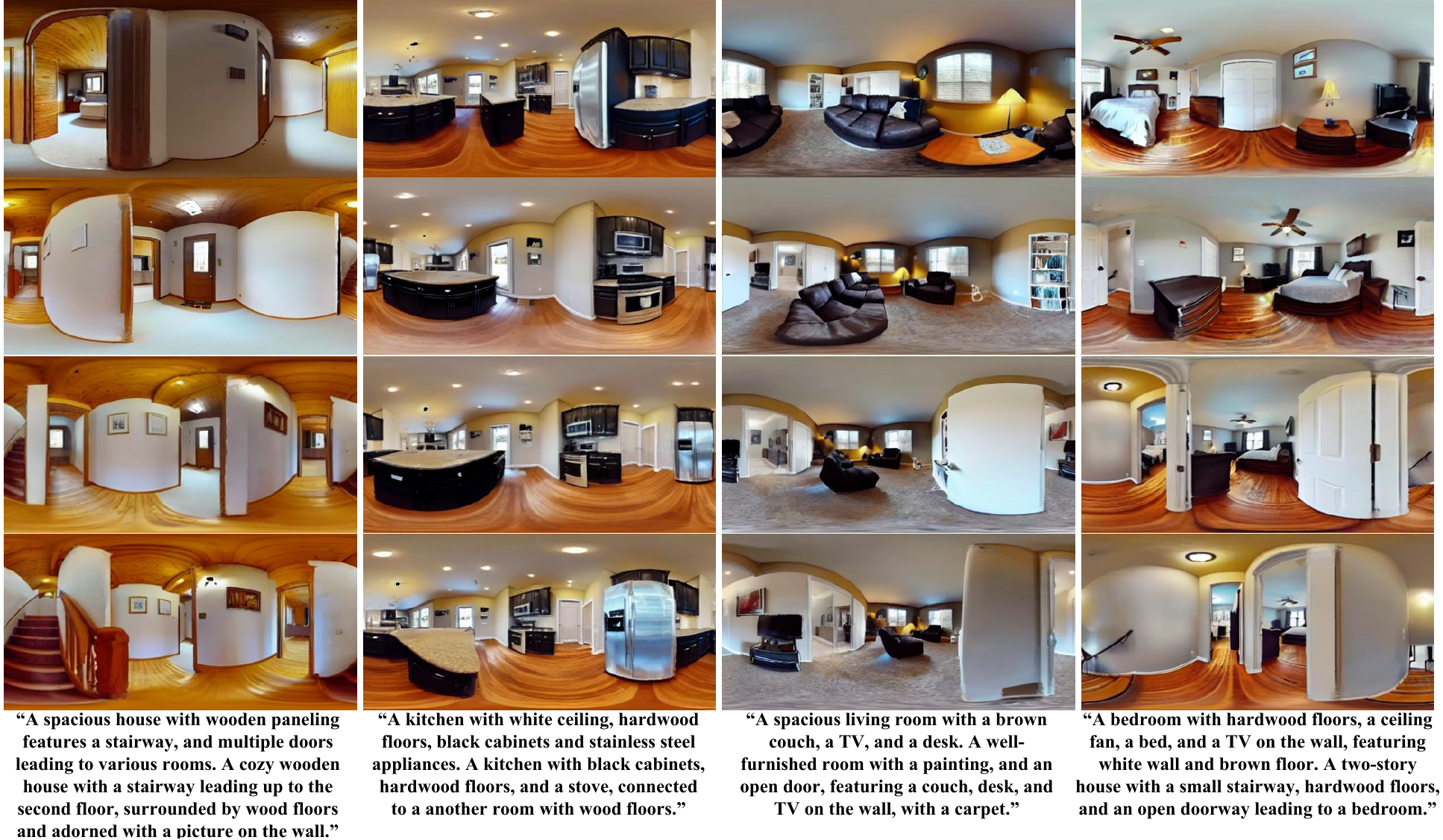}}
        \vspace{-0.1cm}
        \caption{\textbf{Qualitative Results of Text to Panoramic Videos.} DiffPano can generate scalable and consistent panorama videos. }
        \label{fig:pano_videos}
    \end{center}
    \vspace{-0.6cm}
\end{figure}

DiffPano can generate panoramic video frames with large camera pose spans and multi-view consistency based on diverse textual descriptions, thus achieving the effect of text-to-panoramic video, as shown in the Fig.\ref{fig:pano_videos}.

\begin{figure}[ht]
    \begin{center}        
    \centerline{\includegraphics[width=1\linewidth]{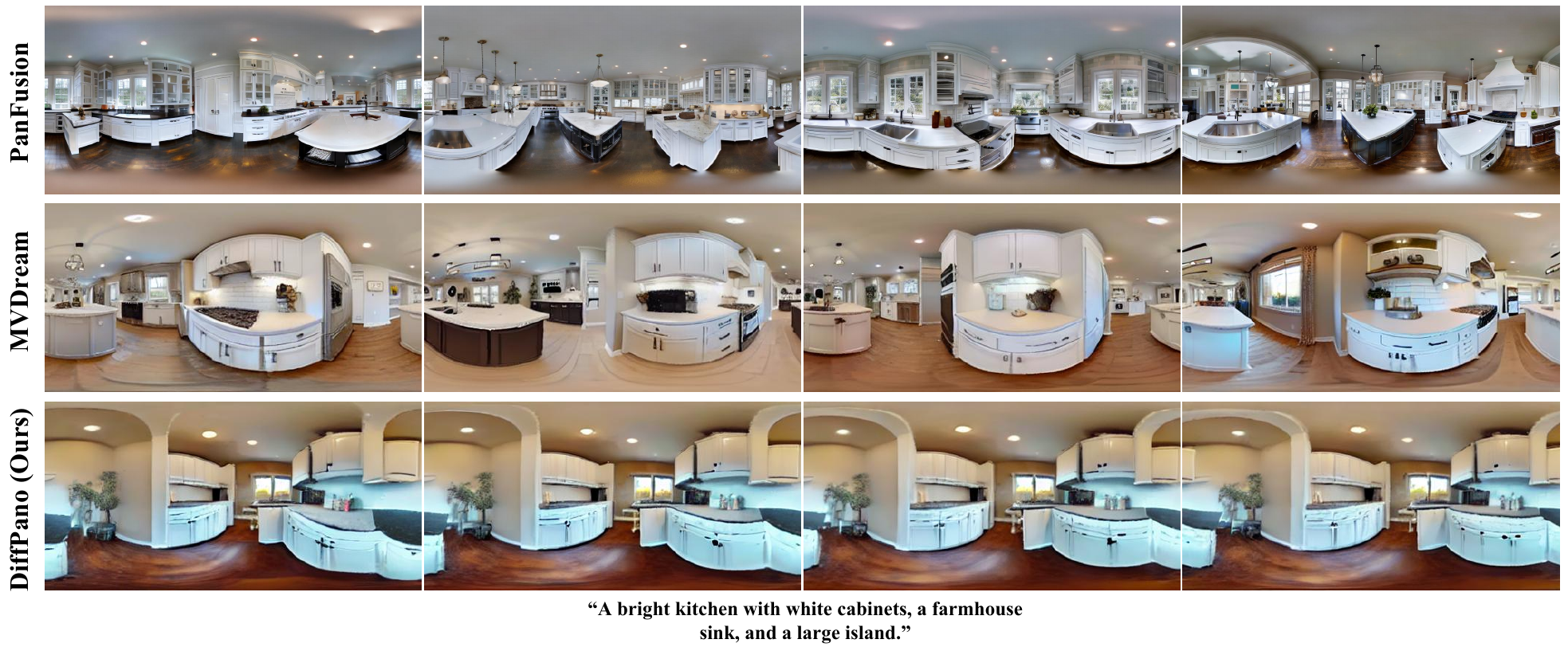}}
        \vspace{-0.1cm}
        \caption{\textbf{Qualitative Comparisons of Text to Panoramic Videos.} Ours vs MVDream vs PanFusion. }
        \label{fig:ablation study}
    \end{center}
    \vspace{-0.6cm}
\end{figure}

\begin{figure}[ht]
    \begin{center}        \centerline{\includegraphics[width=1\linewidth]{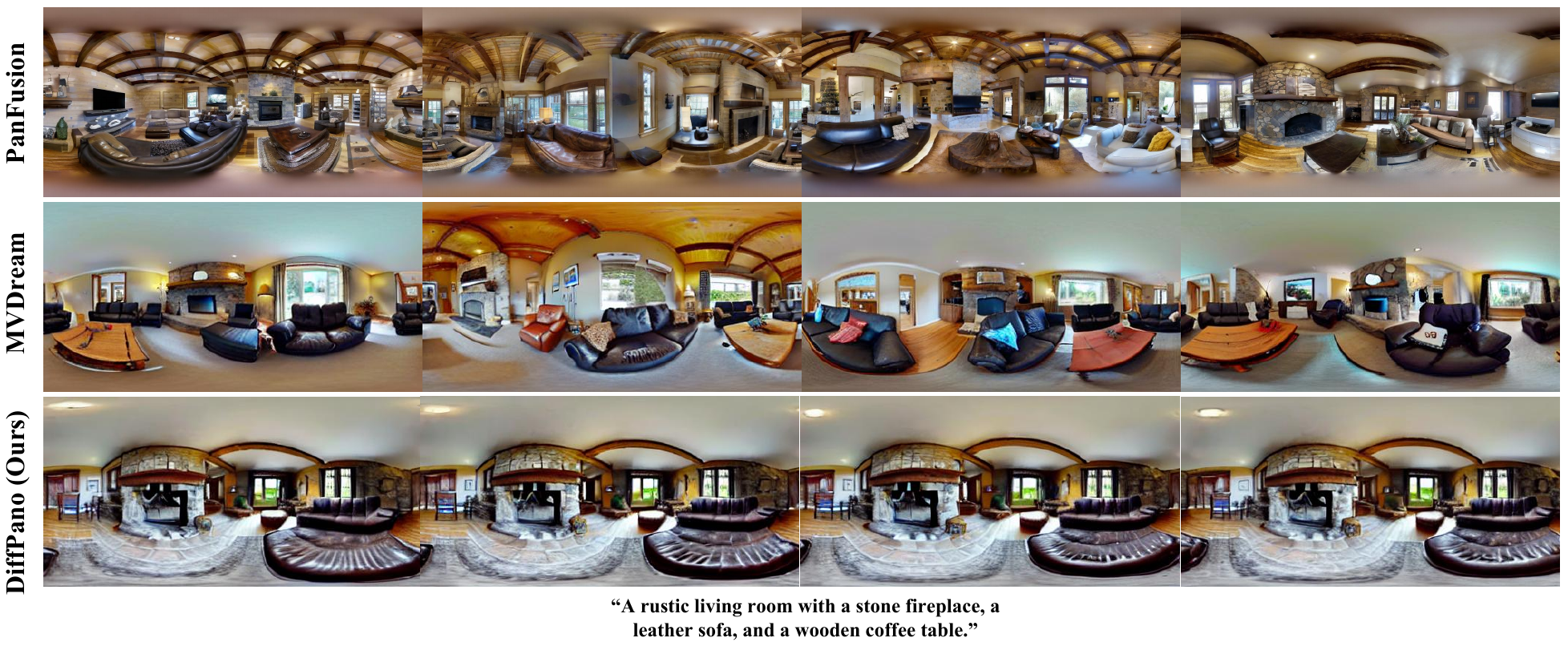}}
        \vspace{-0.1cm}
        \caption{\textbf{Qualitative Comparisons of Text to Panoramic Videos.} Ours vs MVDream vs PanFusion. }
        \label{fig:ablation study2}
    \end{center}
    \vspace{-0.6cm}
\end{figure}

\begin{figure}[ht]
    \begin{center}        \centerline{\includegraphics[width=1\linewidth]{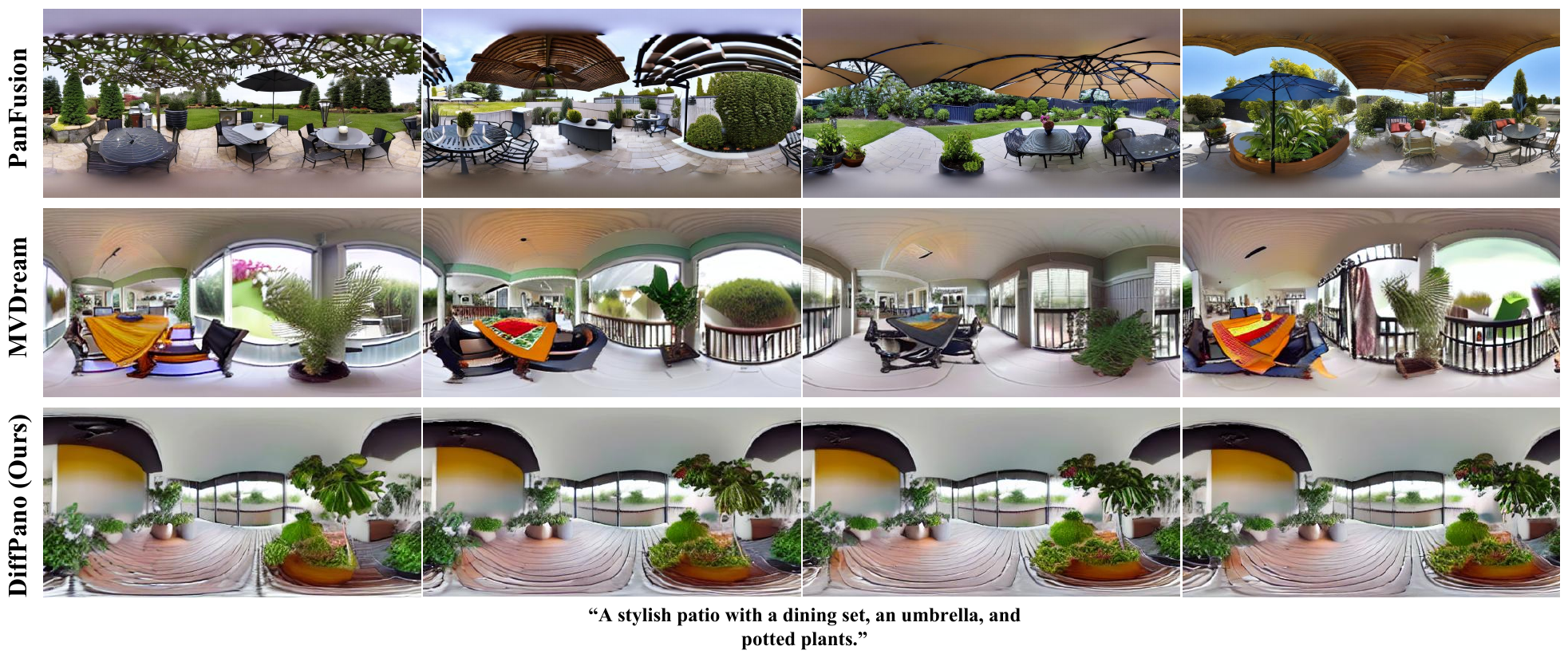}}
        \vspace{-0.1cm}
        \caption{\textbf{Qualitative Comparisons of Text to Panoramic Videos.} Ours vs MVDream vs PanFusion. }
        \label{fig:ablation study3}
    \end{center}
    \vspace{-0.6cm}
\end{figure}

\begin{figure}[ht]
    \begin{center}        \centerline{\includegraphics[width=1\linewidth]{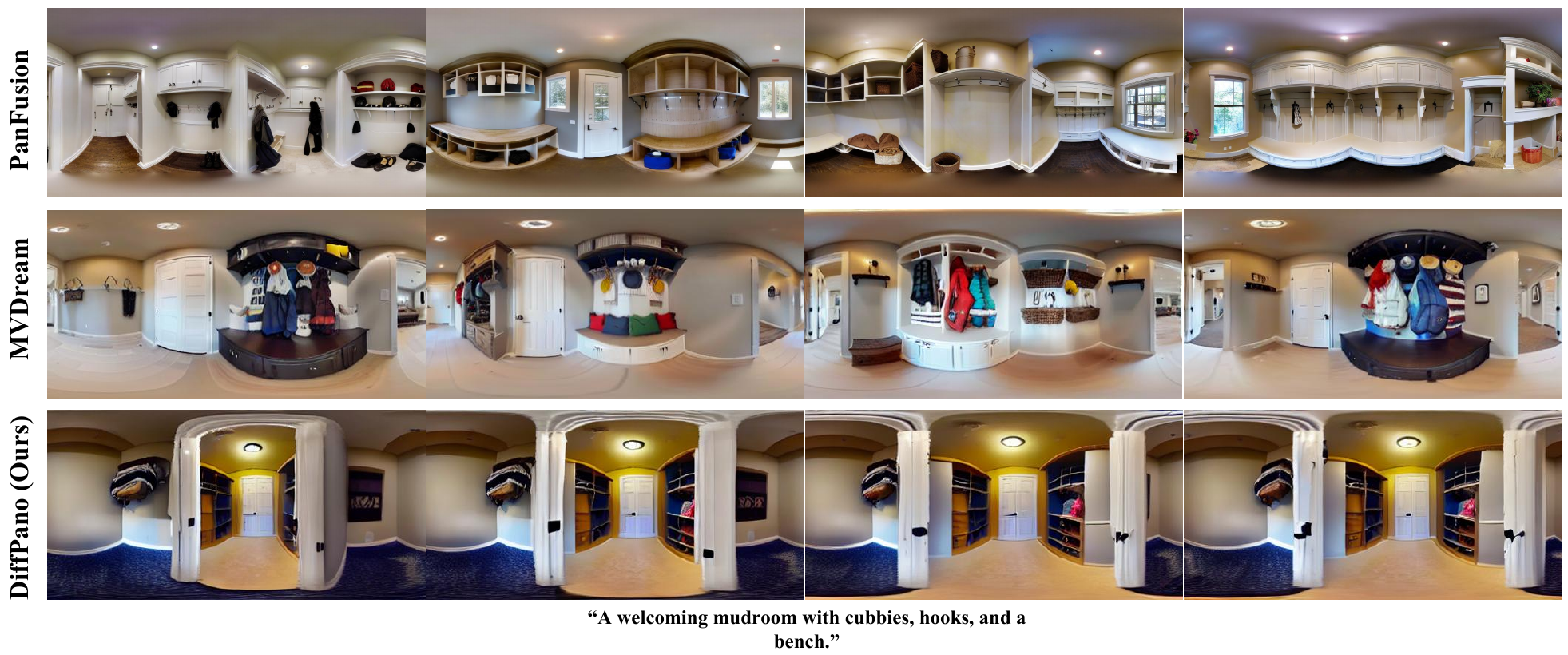}}
        \vspace{-0.1cm}
        \caption{\textbf{Qualitative Comparisons of Text to Panoramic Videos.} Ours vs MVDream vs PanFusion. }
        \label{fig:ablation study4}
    \end{center}
    \vspace{-0.6cm}
\end{figure}

\begin{figure}[ht]
    \begin{center}        \centerline{\includegraphics[width=1\linewidth]{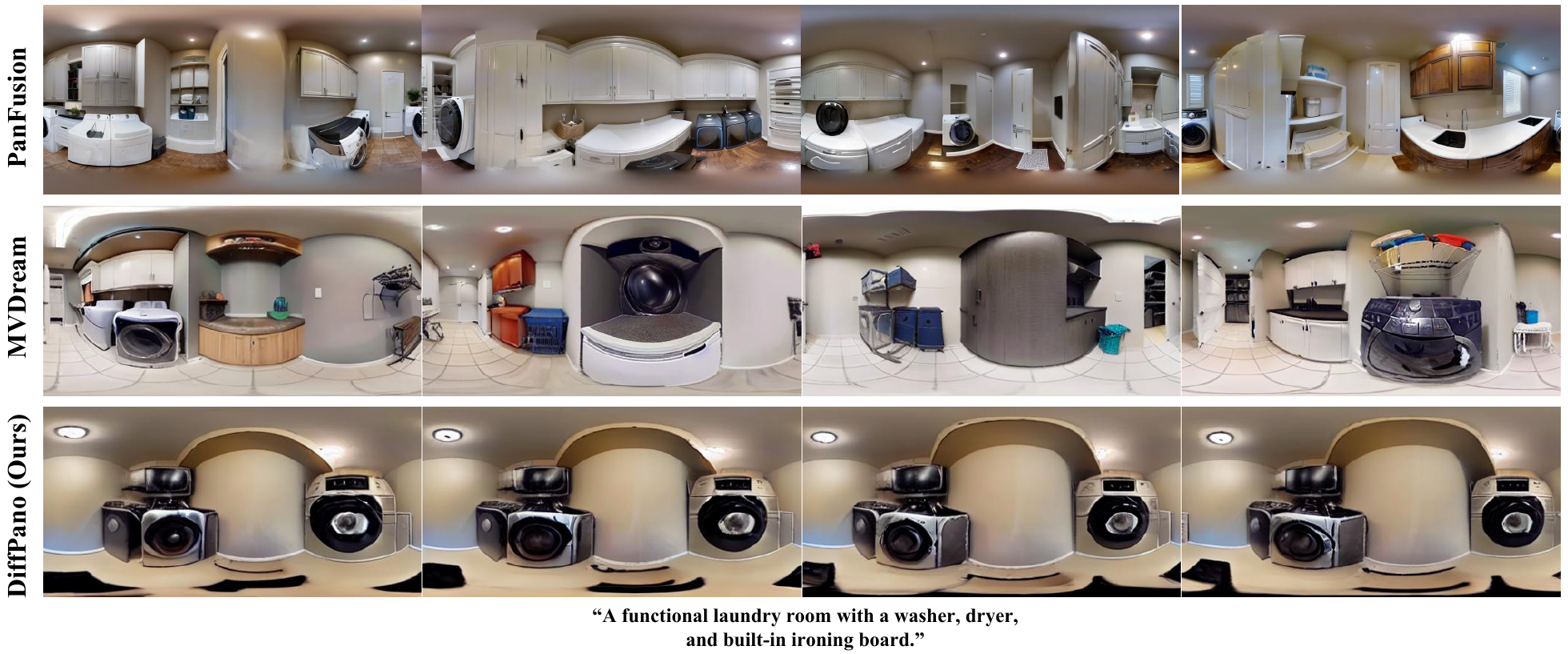}}
        \vspace{-0.1cm}
        \caption{\textbf{Qualitative Comparisons of Text to Panoramic Videos.} Ours vs MVDream vs PanFusion. }
        \label{fig:ablation study5}
    \end{center}
    \vspace{-0.6cm}
\end{figure}

\begin{figure}[ht]
    \begin{center}
    \centerline{\includegraphics[width=1\linewidth]{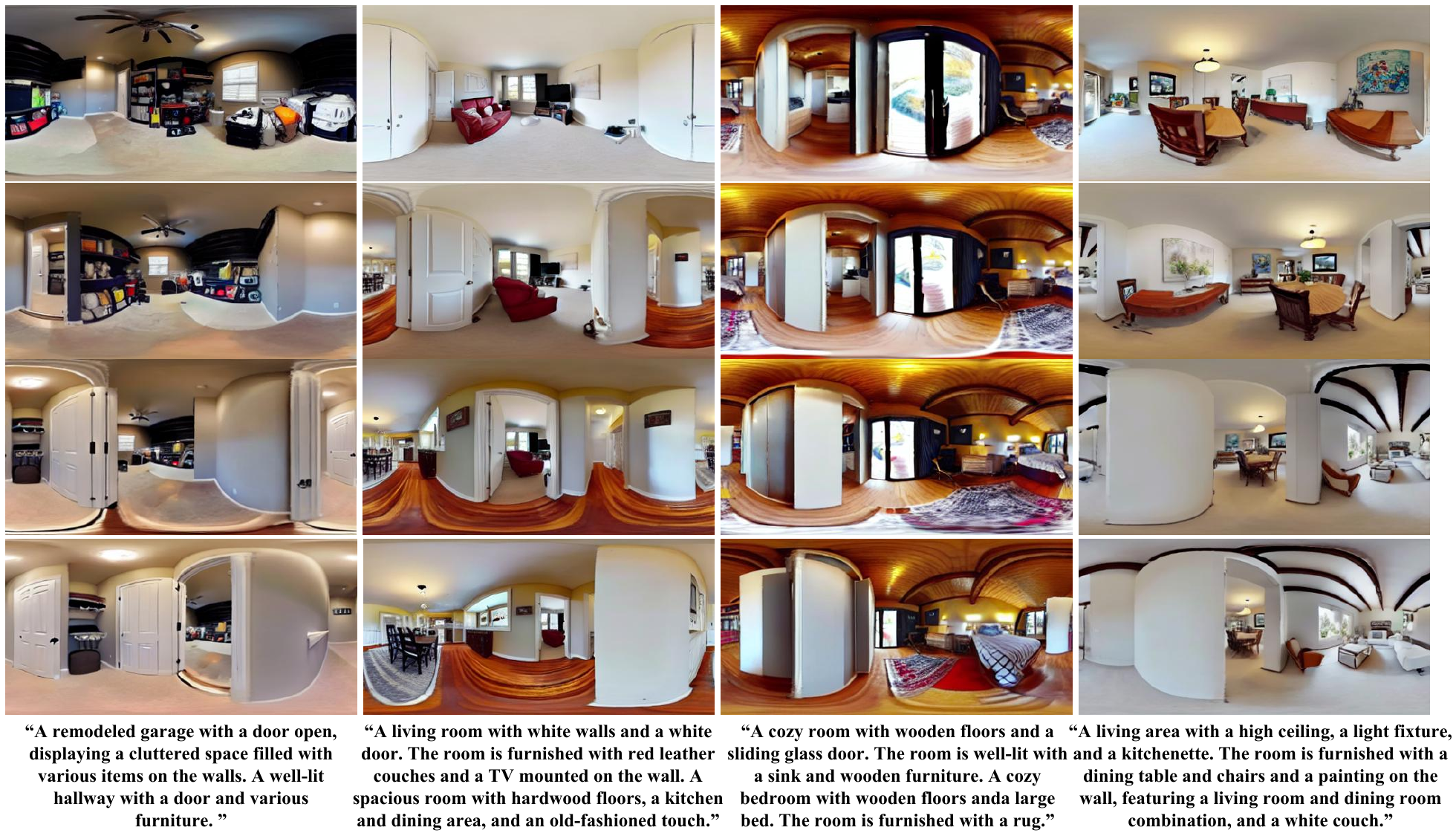}}
        \vspace{-0.1cm}
        \caption{\textbf{Qualitative Results of Text to Panoramic Videos.} DiffPano can generate scalable and consistent panorama videos. }
        \label{fig:pano_videos2}
    \end{center}
    \vspace{-0.6cm}
\end{figure}

\begin{figure}[ht]
    \begin{center}
    \centerline{\includegraphics[width=1\linewidth]{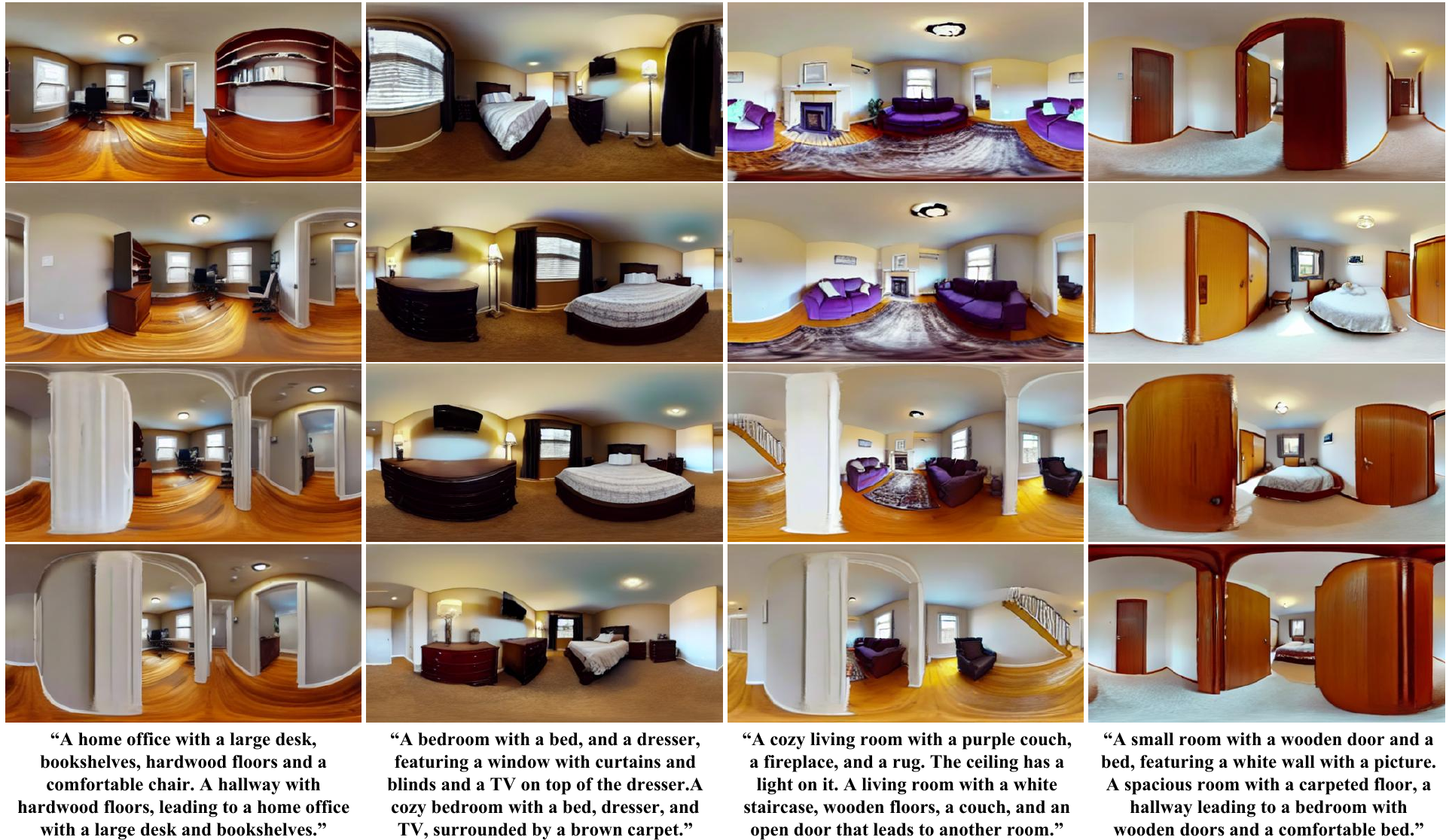}}
        \vspace{-0.1cm}
        \caption{\textbf{Qualitative Results of Text to Panoramic Videos.} DiffPano can generate scalable and consistent panorama videos. }
        \label{fig:pano_videos3}
    \end{center}
    \vspace{-0.6cm}
\end{figure}

\section{Network Architecture and Training Details}
\label{network_detail}
\paragraph{Network Architecture}
% 我们的基于LoRA的单视角全景图生成方法生成的是512x1024分辨率的全景图，在多视角全景图生成方法中生成的是连续多帧的256x512分辨率的全景图，因为一次性生成多帧512x1024分辨率的全景图会消耗大量的计算资源，而且经过我们的实验发现，多帧高分辨率全景图生成需要非常长的训练时间才能够提高生成图像的质量。我们的多视角全景图生成模型的网络结构如表所示。
Our LoRA-based single-view panorama generation model produces panoramas with a resolution of 512$\times$1024. In the multi-view panorama generation approach, we generate continuous frames of panoramas with a resolution of 256$\times$512. Generating multiple frames of 512$\times$1024 resolution panoramas simultaneously would consume a significant amount of computational resources. Moreover, our experiments reveal that generating multi-frame high-resolution panoramas requires an exceptionally long training time to improve the quality of the generated images. The network architecture of our multi-view panorama generation model is shown in Table \ref{network_architecture_1} and Table\ref{network_architecture_2}.

\begin{table}
  \caption{Network architecture of DiffPano-1}
  \label{network_architecture_1}
  \centering
  \begin{tabular}{lccc}
    \toprule
     & Layer     & Output     & Additional
     Inputs  \\
    \midrule
(1) & Latent Map  & $4\times32\times64$ &     \\
(2) & Conv. & $320\times32\times64$ &     \\
    \midrule
    \multicolumn{4}{c}{CrossAttnDownBlock1} \\
    \midrule
(3) & ResBlock & $320\times32\times64$ & Time emb.  \\
(4) & AttnBlock & $320\times32\times64$ & Prompt emb.  \\
(5) & ResBlock & $320\times32\times64$ & Time emb.  \\
(6) & AttnBlock & $320\times32\times64$ & Prompt emb.  \\
(7) & DownSampler & $320\times16\times32$ & \\
    \midrule
    \multicolumn{4}{c}{CrossAttnDownBlock2} \\
    \midrule
(8) & ResBlock & $640\times16\times32$ & Time emb.  \\
(9) & AttnBlock & $640\times16\times32$ & Prompt emb.  \\
(10) & ResBlock & $640\times16\times32$ & Time emb.  \\
(11) & AttnBlock & $640\times16\times32$ & Prompt emb.  \\
(12) & DownSampler & $640\times8\times16$ & \\
    \midrule
    \multicolumn{4}{c}{CrossAttnDownBlock3} \\
    \midrule
(13) & ResBlock & $1280\times8\times16$ & Time emb.  \\
(14) & AttnBlock & $1280\times8\times16$ & Prompt emb.  \\
(15) & ResBlock & $1280\times8\times16$ & Time emb.  \\
(16) & AttnBlock & $1280\times8\times16$ & Prompt emb.  \\
(17) & DownSampler & $1280\times4\times8$ & \\
    \midrule
    \multicolumn{4}{c}{DownBlock} \\
    \midrule
(18) & ResBlock & $1280\times4\times8$ & Time emb.  \\
(19) & ResBlock & $1280\times4\times8$ & Time emb.  \\
    \midrule
    \multicolumn{4}{c}{MidBlock} \\
    \midrule
(20) & ResBlock & $1280\times4\times8$ & Time emb.  \\
\bf{(21)} & \bf{EAModule} & $\bf{1280}\times\bf{4}\times\bf{8}$ &  \\
(22) & AttnBlock & $1280\times4\times8$ & Prompt emb.  \\
(23) & ResBlock & $1280\times4\times8$ & Time emb.  \\
    \midrule
    \multicolumn{4}{c}{UpBlock} \\
    \midrule
(24) & ResBlock & $1280\times4\times8$ & (19), Time emb. \\
(25) & ResBlock & $1280\times4\times8$ & (18), Time emb. \\
(26) & ResBlock & $1280\times4\times8$ & (17), Time emb. \\
\bf{(27)} & \bf{EAModule} & $\bf{1280}\times\bf{4}\times\bf{8}$ &  \\
(28) & UpSampler & $1280\times8\times16$ & \\
    \midrule
    \multicolumn{4}{c}{CrossAttnUpBlock1} \\
    \midrule
(29) & ResBlock & $1280\times8\times16$ & (16), Time emb.\\
(30) & AttnBlock & $1280\times8\times16$ & Prompt emb.  \\
(31) & ResBlock & $1280\times8\times16$ & (14), Time emb.\\
(32) & AttnBlock & $1280\times8\times16$ & Prompt emb.  \\
(33) & ResBlock & $1280\times8\times16$ & (12), Time emb.\\
(34) & AttnBlock & $1280\times8\times16$ & Prompt emb.  \\
\bf{(35)} & \bf{EAModule} & $\bf{1280}\times\bf{8}\times\bf{16}$ &  \\
(36) & UpSampler & $1280\times16\times32$ & \\
    \bottomrule
  \end{tabular}
\end{table}

\begin{table}
  \caption{Network architecture of DiffPano-2}
  \label{network_architecture_2}
  \centering
  \begin{tabular}{lccc}
    \toprule
     & Layer     & Output     & Additional
     Inputs  \\
    \midrule
    \multicolumn{4}{c}{CrossAttnUpBlock2} \\
    \midrule
(37) & ResBlock & $640\times16\times32$ & (11), Time emb.\\
(38) & AttnBlock & $640\times16\times32$ & Prompt emb.  \\
(39) & ResBlock & $640\times16\times32$ & (9), Time emb.\\
(40) & AttnBlock & $640\times16\times32$ & Prompt emb.  \\
(41) & ResBlock & $640\times16\times32$ & (7), Time emb.\\
(42) & AttnBlock & $640\times16\times32$ & Prompt emb.  \\
\bf{(43)} & \bf{EAModule} & $\bf{640}\times\bf{16}\times\bf{32}$ &  \\
(44) & UpSampler & $640\times32\times64$ & \\
    \midrule
    \multicolumn{4}{c}{CrossAttnUpBlock3} \\
    \midrule
(45) & ResBlock & $320\times32\times64$ & (6), Time emb.\\
(46) & AttnBlock & $320\times32\times64$ & Prompt emb.  \\
(47) & ResBlock & $320\times32\times64$ & (4), Time emb.\\
(48) & AttnBlock & $320\times32\times64$ & Prompt emb.  \\
(49) & ResBlock & $320\times32\times64$ & (2), Time emb.\\
(50) & AttnBlock & $320\times32\times64$ & Prompt emb.  \\
\bf{(51)} & \bf{EAModule} & $\bf{320}\times\bf{32}\times\bf{64}$ &  \\
    \midrule
(52) &  GroupNorm & $320\times32\times64$ &   \\
(53) &  SiLU & $320\times32\times64$ &   \\
(54) &  Conv. & $4\times32\times64$ &   \\
    \bottomrule
  \end{tabular}
\end{table}

\paragraph{Training Details}
% 对于多视角全景图像生成，我们在8张80G A100 GPU上以batchsize为1，学习率为1e-5训练(每张卡占了大约50%显存)，在两阶段训练过程中分别训练100轮(共需约5天时间)。
We fine-tuned the Stable Diffusion v1.5 model using the LoRA method for single-view pano-based synthesis. The training was conducted on 6 A100 GPUs with 80GB memory for 100 epochs (approximately 6.5 hours), with a learning rate of 1e-4 and a batch size of 6.
For multi-view panoramas generation, we conducted training on 8 80G A100 GPUs with a batch size of 1 and a learning rate of 1e-5. Each GPU utilized approximately 50\% of its memory. The two-stage training process involved 100 epochs for each stage, with a total training time of approximately 5 days.
%%%%%%%%%%%%%%%%%%%%%%%%%%%%%%%%%%%%%%%%%%%%%%%%%%%%%%%%%%%%

\section{Societal Impact}\label{sec:supp.impact}
Since our method can achieve scalable, consistent, and diverse multi-view panoramas, it has many potential applications, such as unlimited room roaming in VR, interior design preview, embodied intelligent robot exploration, etc.

\end{document}